\documentclass{article}

\usepackage[numbers]{natbib}

\usepackage{graphicx}
\usepackage{blindtext}
\usepackage{floatrow}
\usepackage{booktabs}
\usepackage{multirow}

\usepackage[preprint]{neurips_2025}

\usepackage[utf8]{inputenc} 
\usepackage[T1]{fontenc}    
\usepackage[bookmarks=false]{hyperref}       
\usepackage{url}            
\usepackage{booktabs}       
\usepackage{amsfonts}       
\usepackage{nicefrac}       
\usepackage{microtype}      
\usepackage{xcolor}         
\usepackage{amsmath,amssymb}
\usepackage[most]{tcolorbox}
\usepackage{listings}
\tcbuselibrary{listingsutf8}
\usepackage{placeins}

\title{Hallucinate, Ground, Repeat: A Framework for Generalized Visual Relationship Detection}

\author{%
  Shanmukha Vellamcheti, Sanjoy Kundu, Sathyanarayanan N. Aakur \\
  CSSE Department, Auburn University\\
Auburn, Alabama, USA 36849\\
  \texttt{\{szv0080,szk0266,san0028\}@auburn.edu} \\
}

\begin{document}

\maketitle

\begin{abstract}
Understanding relationships between objects is central to visual intelligence, with applications in embodied AI, assistive systems, and scene understanding. Yet, most visual relationship detection (VRD) models rely on a fixed predicate set, limiting their generalization to novel interactions. A key challenge is the inability to visually ground semantically plausible, but unannotated, relationships hypothesized from external knowledge.
This work introduces an iterative visual grounding framework that leverages large language models (LLMs) as structured relational priors. Inspired by expectation-maximization (EM), our method alternates between generating candidate scene graphs from detected objects using an LLM (expectation) and training a visual model to align these hypotheses with perceptual evidence (maximization). This process bootstraps relational understanding beyond annotated data and enables generalization to unseen predicates. 
Additionally, we introduce a new benchmark for open-world VRD on Visual Genome with 21 held-out predicates and evaluate under three settings: seen, unseen, and mixed. Our model outperforms LLM-only, few-shot, and debiased baselines, achieving mean recall (mR@50) of 15.9, 13.1, and 11.7 on predicate classification on these three sets. These results highlight the promise of grounded LLM priors for scalable open-world visual understanding. 

\end{abstract}

\section{Introduction}
\begin{figure*}[t]
    \centering
    \includegraphics[width=0.9\linewidth]{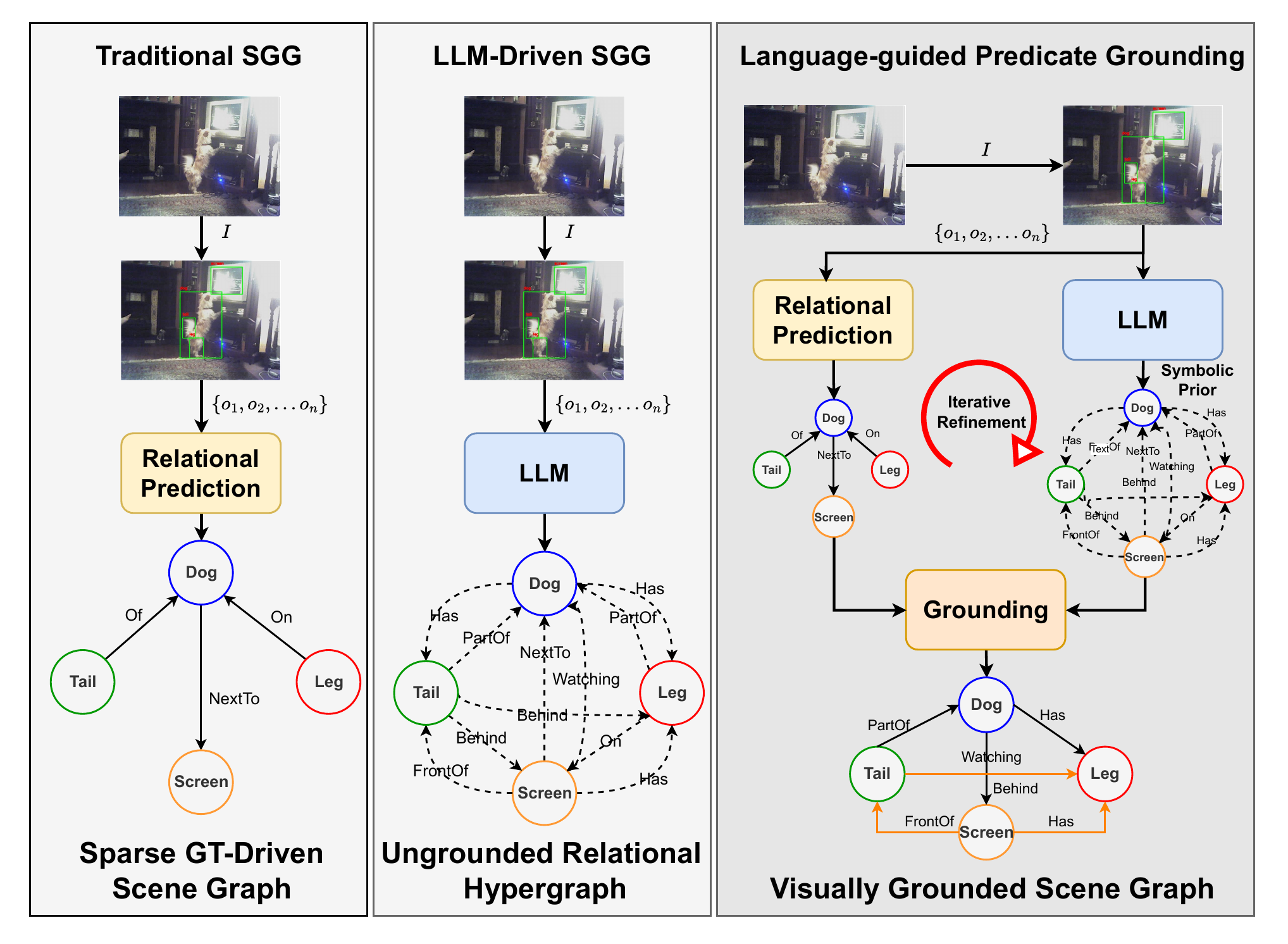}
    \caption{\textbf{Overview}. (Left) Traditional SGG relies on sparse annotations. (Middle) LLMs can hallucinate relationships, but are ungrounded, multi-relational hypergraphs. (Right) Our method leverages these hallucinations and grounds them through iterative visual alignment. 
    }
    \label{fig:overview}
\end{figure*}

Understanding object relationships is central to high-level visual reasoning. Visual Relationship Detection (VRD), which encodes interactions as (subject, predicate, object) triplets, underpins Scene Graph Generation (SGG), providing structured representations useful for embodied navigation, assistive perception, and open-domain image understanding. Yet most existing SGG models operate under a closed-world assumption, relying on a fixed predicate vocabulary and dense human supervision. As illustrated in Figure~\ref{fig:overview}, such models are constrained by sparse, saliency-biased annotations that capture a small subset of valid interactions, limiting generalization to novel or rare relationships. Relevant information from the non-salient and background areas is left unused. It can be leveraged to capture the underlying relational structure between all objects in the scene, even if they are not of interest in that image's context. Such information, if objectively sampled, can enhance generalization to unseen and potentially unknown relationships. 

To address this, we propose a shift from annotation-driven learning to a prior-driven framework. As shown in Figure~\ref{fig:overview} (middle), large language models (LLMs) can hallucinate symbolic graphs from detected object labels to produce a rich, overcomplete relational hypergraph that encodes commonsense and co-occurrence priors. While not grounded in visual input, these symbolic hypotheses form a structured prior that can be selectively aligned with image evidence. Our method (Figure~\ref{fig:overview}, right) frames this as an EM-style optimization: LLMs propose candidate triplets, and a visual grounding model iteratively filters and refines them based on perceptual support. This formulation enables generalized VRD, scaling beyond annotated labels to recognize seen and unseen predicates through symbolic guidance and grounded refinement. 
Our approach, EM-Grounding, decouples visual relationship prediction into two phases: symbolic hallucination and perceptual grounding. Given object detections from an image, an LLM generates a multi-relational prior, proposing multiple predicates per object pair. These form a symbolic hypergraph that is pruned through an iterative visual alignment model trained solely on LLM-generated triplets. Through successive refinement cycles, our model recovers semantically valid, visually grounded relationships, even without access to labeled edge annotations. This separation between reasoning and grounding enables scalable training with minimal supervision.

In this paper, we make the following \textbf{contributions}:
(i) we propose EM-Grounding, a novel weakly supervised framework that treats large language models as symbolic priors and grounds them through perceptual alignment in an iterative EM-style process;
(ii) introduce a semantic relational hypergraph formulation that captures multiple plausible predicates per object pair and design a visual model to resolve ambiguity through visual grounding;
(iii) we demonstrate that EM-Grounding enables generalization to unseen predicates and scales relationship recognition beyond human annotations, outperforming all weakly supervised and few-shot baselines; and
(iv) we present a comprehensive benchmark on Visual Genome with held-out predicates and mixed evaluation settings. 

Our results (Section~\ref{sec:results}) show that EM-Grounding improves recall on unseen predicates and achieves state-of-the-art performance under weak supervision, rivaling supervised models trained on vastly more data. These gains persist across different scene graph generation tasks, highlighting the potential of symbolic priors for enabling generalizable, scalable visual reasoning. 

\section{Prior Works}
\textbf{Scene graph generation (SGG)} was introduced by \citet{johnson2015image} to support high-level image retrieval via structured representations of objects, attributes, and relationships. Since then, scene graphs have become central to a range of downstream tasks, including navigation~\cite{9812179,Singh_2023_ICCV}, visual question answering~\cite{nuthalapati2021lightweight}, image manipulation~\cite{dhamo2020semantic}, and captioning~\cite{nguyen2021defense}. Large-scale benchmarks like Visual Genome~\cite{krishna2017visual} have catalyzed progress in SGG frameworks, which typically begin with object detection followed by pairwise interaction modeling to capture relational context~\cite{cong2023reltr,kundu2023ggt,10.1007/978-3-031-19836-6_24,xu2017scene,yang2018graph,zellers2018neural}. While performing well on common predicates, these models struggle with rare relationships due to the severe long-tail distribution inherent in SGG benchmarks.

\textbf{Addressing long-tail bias.} Naïve reweighting of predicate frequencies offers marginal gains but often reduces recall for common classes. To mitigate this, many works introduce explicit bias-handling techniques. Some integrate external knowledge bases~\cite{chen2019knowledge,Tang_2019_CVPR,zareian2020bridging}, while others leverage causal or counterfactual reasoning~\cite{Tang_2020_CVPR} or energy-based models~\cite{Suhail_2021_CVPR}. Hierarchical and cognitively inspired strategies include CogTree~\cite{yu2020cogtree}, RU-Net~\cite{lin2022ru}, BGNN~\cite{li2021bipartite}, IETrans~\cite{zhang2022fine}, and HiKER-SGG~\cite{zhang2024hiker}. Meanwhile, tailored loss functions improve supervision for rare predicates by rebalancing based on predicate context (PCPL~\cite{yan2020pcpl}, FGPL~\cite{lyu2022fine}, A-FGPL~\cite{lyu2023adaptive}), correcting label noise (NICE~\cite{li2022devil}), or leveraging predicate-level distributions (PDPL~\cite{li2022ppdl}, DLFE~\cite{chiou2021recovering}). These approaches reflect a growing recognition that overcoming annotation bias and sparsity requires structured priors, auxiliary signals, and generalizable representations beyond conventional closed-world assumptions. While bias-mitigation techniques improve recognition of rare predicates, they remain constrained by the fixed predicate set and closed-world assumptions of existing benchmarks. 

\textbf{Generalized visual understanding} requires models to recognize both seen and unseen relationships, extending beyond predefined semantics. Dubbed `open-world learning,'' there have been numerous efforts in image classification~\cite{bendale2015towards, shu2018unseen}, zero-shot recognition with vision-language models (VLMs) like CLIP~\cite{clip}, BLIP~\cite{li2022blip}, and their extensions to video~\cite{brattoli2020rethinking,xu2021videoclip}. In object detection, Open World DETR~\cite{dong_open_2022} and open-vocabulary detectors~\cite{du_learning_2022,gu_open-vocabulary_2022} use VLMs via prompting and distillation, though issues like class bias persist~\cite{xi_umb_2024}. Parallel work in open-vocabulary activity recognition~\cite{chatterjee2023opening,wu2024open} and open-world event understanding~\cite{aakur2022knowledge, kundu2024discovering, kundu2025probres} explores temporal reasoning. Recently, open-world scene graph generation (SGG) has gained interest: CaCao~\cite{Yu_2023_ICCV} introduces visually-prompted LLMs for zero-shot predicate generation, while others tackle open-vocabulary~\cite{chen2024expanding, zhao2023less, Zhong_2021_ICCV} and panoptic~\cite{zhou2024openpsg} SGG. Open-world SGG has also been applied to tasks like object navigation~\cite{loo2024open}. 

While these approaches expand the scope of open-world understanding, they often rely on visual-text alignment or retrieval, without structured reasoning over relational hypotheses. In contrast, our work leverages large language models not merely as classifiers, but as symbolic priors for structured relationship grounding, building on recent advances in using \textbf{LLMs as knowledge sources}. 
For example, Large Language Models (LLMs) have emerged as powerful implicit knowledge sources, capable of encoding and retrieving structured relational knowledge. Early work demonstrated this capacity in pretrained transformers like BERT~\cite{petroni2019language}, while recent efforts explore hybrid systems that combine LLMs with explicit knowledge graphs~\cite{pan2023large, wang2023knowledgpt}. Studies also examine the scope and factual accuracy of LLM knowledge~\cite{hu2024towards} and propose KG-guided prompting to improve grounding~\cite{zhang2024knowgpt}. 

\section{Hallucinations, Grounding, and Visual Alignment}

\textbf{Visual relationship detection (VRD)} aims to identify semantic interactions between objects in an image, typically expressed as triplets of the form $(s, p, o)$, where $s$ and $o$ denote subject and object entities, and $p$ is a predicate describing their interaction (e.g., \textit{(person, holding, cup)}). This task forms the backbone of scene graph generation (SGG), where the goal is to construct a structured representation of the scene by predicting all valid relational triplets between detected objects. 

\begin{figure*}[t]
    \begin{tabular}{ccc}
    \includegraphics[width=0.25\linewidth]{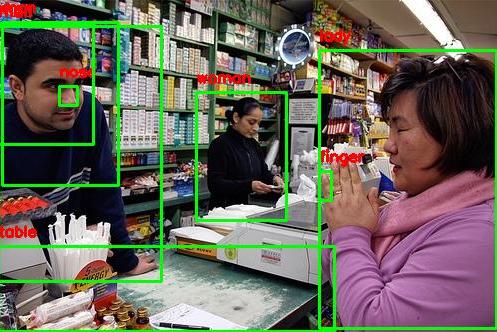} &  
        \includegraphics[width=0.25\linewidth]{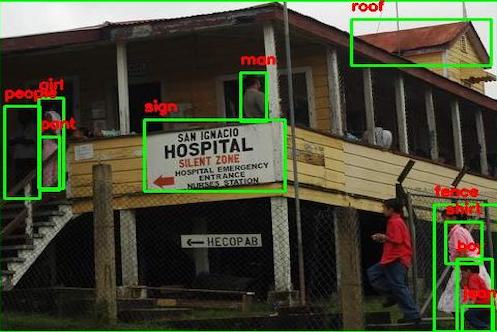} & 
         \includegraphics[width=0.25\linewidth]{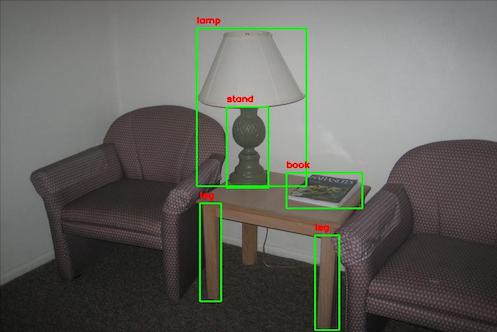} \\
        \includegraphics[width=0.3\linewidth]{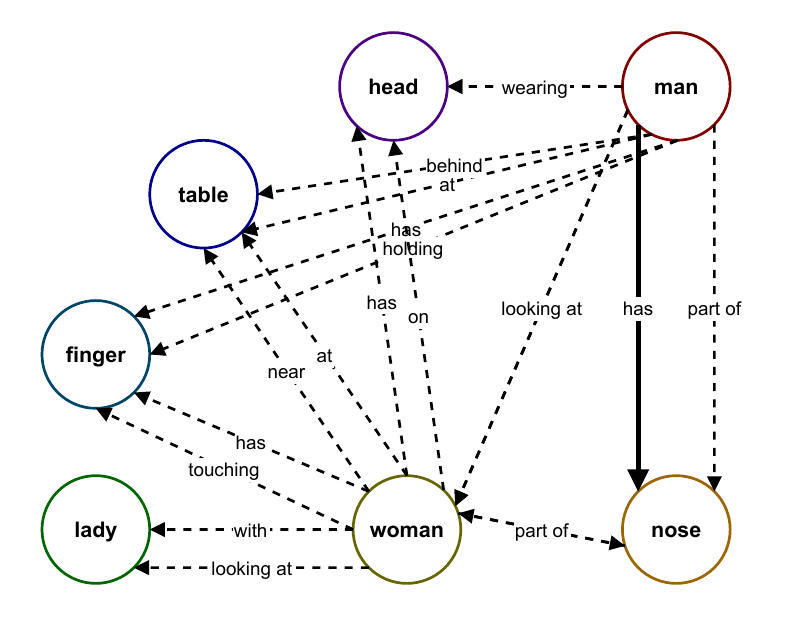} &  
        \includegraphics[width=0.3\linewidth]{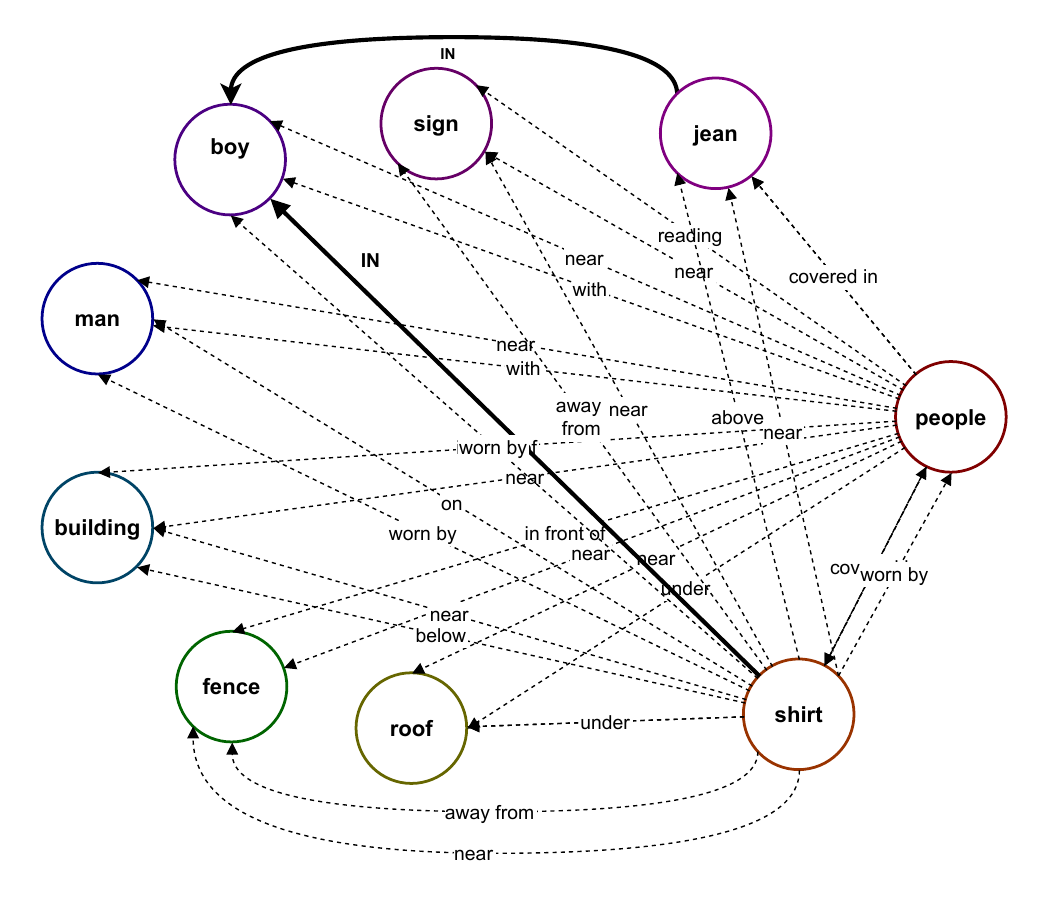} & 
         \includegraphics[width=0.3\linewidth]{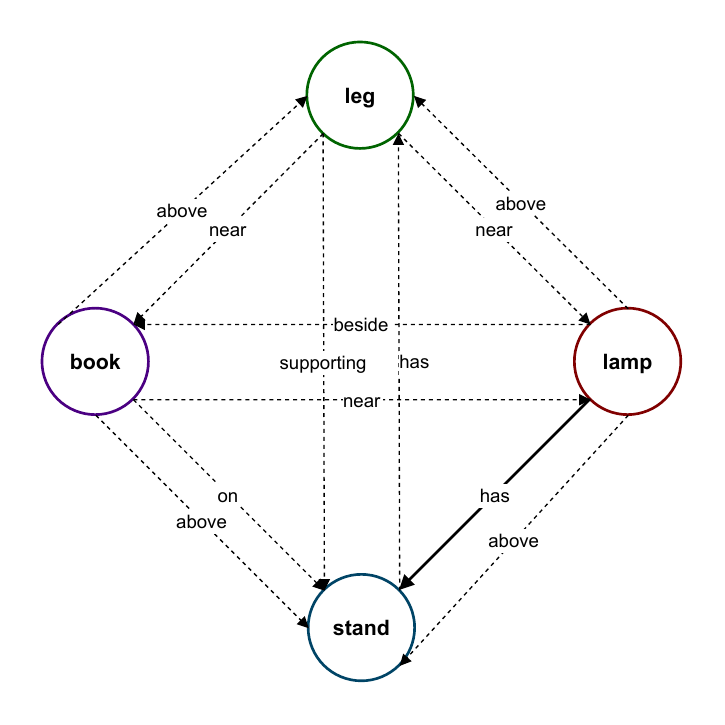} \\
         (a) & (b) & (c) \\
    \end{tabular}
    \caption{\textbf{Sparse annotations} in Visual Genome. 
    Top: Reference images. Bottom: corresponding relational hypergraphs, where solid edges are ground-truth relationships and dashed edges are plausible but unannotated ones. 
    Annotations often ignore valid relations such as ``lamp on stand.'' 
    }
    \label{fig:vg-missing}
\end{figure*}
While datasets like Visual Genome~\cite{krishna2017visual} have fueled progress in VRD and SGG, their human-annotated triplets are inherently incomplete. Cognitive constraints, such as attentional saliency~\cite{itti2001computational,wolfe1994guided}, limited annotation time~\cite{maule2002effects}, and task framing~\cite{folk1992involuntary,simons1999gorillas}, lead annotators to focus on a small subset of relationships, often those that are spatially prominent or contextually salient (e.g., \textit{on}, \textit{next to}, \textit{wearing}). As a result, many semantically valid relationships go unannotated, especially in visually dense scenes. 
 
Figure~\ref{fig:vg-missing} shows examples from the Visual Genome dataset where the annotations are sparse and focused on salient, frequent relationships while ignoring additional information in the scene. 

This sparsity restricts the learnable semantics and hinders generalization by biasing models toward frequent relationships.

We treat large language models (LLMs) as structured priors over scene semantics to address annotation sparsity. Given detected objects, an LLM can hallucinate plausible triplets based on co-occurrence, commonsense, and world knowledge, e.g., \textit{(person, drinking from, bottle)} or \textit{(bottle, on, table)}, even if such relationships are unannotated. These hallucinations reflect semantic plausibility rather than visual evidence, but we argue they can, and should, be grounded in the image. Human omission often stems from cognitive salience, not the absence of a relation. LLM-generated triplets thus define a rich hypothesis space that can be grounded through visual models. 
We formalize the LLM-hallucinated triplets as latent variables in a weakly supervised learning framework. Since the LLM often predicts multiple plausible predicates for a single object pair $(s, o)$, the resulting structure is a multi-relational hypergraph, i.e., a symbolic prior that captures semantically valid but unverified interactions. Visual grounding aims to disambiguate this overcomplete set by aligning triplets with visual evidence. 

To this end, we adopt an expectation-maximization (EM) formulation. In the \textit{expectation step}, the LLM produces a symbolic prior $P_{\text{prior}}(s, p, o \mid O)$ over potential triplets given detected objects. In the \textit{maximization step}, we train a generative visual model to align these symbolic hypotheses with perceptual input. Triplets consistent with the image are reinforced, while unsupported ones are left untouched. The model filters noisy priors through this iterative process and progressively learns relational semantics beyond human-annotated semantics. 

This is defined as the objective for

\begin{equation}
\max_{\theta} ; \mathbb{E}{I, O} \left[ \log \sum{\mathcal{T}^* \subseteq \mathcal{T}} P_{\theta}(\mathcal{T}^* \mid I) \cdot P_{\text{prior}}(\mathcal{T}^* \mid O) \right]
\label{eqn:grounding_opt}
\end{equation}

where $I$ is the input image, $O$ the set of detected objects, $\{\mathcal{T}^*, \hat{\mathcal{T}}\} \in \mathcal{T}$ is the space of all candidate triplets.   
$P_{\text{prior}}$ is the symbolic prior derived from LLM outputs, and $P_\theta$ is the visual model estimating the likelihood that a triplet is grounded in the image. While we refer to $P_{\text{prior}}$ for notational consistency, the triplets $\hat{\mathcal{T}}$ are deterministically produced from ranked LLM outputs rather than sampled probabilistically. 
This formulation captures the core objective of our framework: to recover the subset $\mathcal{T}^* \subseteq \hat{\mathcal{T}}$ of triplets that are both semantically coherent and visually supported. Since exact inference is intractable, our method approximates this objective through an iterative EM-style process that grounds structured priors in perceptual evidence.
We define this process of validating symbolic hypotheses through visual input as \textit{grounding}, and refer to the learned mapping between symbolic triplets and image features as \textit{visual alignment}. Rather than using LLM outputs as supervision, we use them to define a structured latent prior that can be grounded and refined through perception.

\section{Our Method: EM-Style Predicate Grounding}
\textbf{Overview.} 
Given an input image $I$ and a set of detected objects $O = \{o_1, \dots, o_n\}$, the goal of visual relationship detection is to predict a set of relational triplets $\mathcal{T} = \{(s, p, o)\}$, where $(s, o) \in O \times O$ and $p$ is a semantic predicate. We assume that the true set of grounded relationships $\mathcal{T}^* \subseteq \mathcal{T}$ is only partially observed in existing datasets due to annotation sparsity. To address this, we incorporate a symbolic prior $P_{\text{prior}}(s, p, o \mid O)$ derived from a large language model (LLM), which defines a distribution over plausible but ungrounded triplets. However, the symbolic prior derived from an LLM produces a multi-relational structure, proposing multiple plausible predicates per object pair. Our task is to identify which of these prior triplets are visually supported by learning a visual grounding model $P_\theta(s, p, o \mid I)$. 
As formalized in Equation~\ref{eqn:grounding_opt}, we approximate the latent alignment between perceptual evidence and symbolic priors through an iterative procedure, alternating between hallucinating relational hypotheses and refining them via visual grounding.

\subsection{Expectation Step: Triplet Hypothesis Generation via Language Priors}

To instantiate the symbolic prior $P_{\text{prior}}(s, p, o \mid O)$ introduced in Equation~\ref{eqn:grounding_opt}, we use a large language model (LLM), specifically GPT-4o, to hallucinate plausible relational triplets based solely on the object categories detected in each image. For each image $I$, we extract a set of object detections $O = \{o_1, \dots, o_n\}$ using a standard object detector trained on the 150 object classes defined in the Visual Genome dataset. No additional context—such as object position, number of instances, spatial layout, or attributes—is provided to the LLM. This ensures that the hallucinated triplets reflect only prior knowledge and semantic plausibility, independent of the visual input. 
The LLM is prompted with the full list of object categories, all possible ordered object pairs from the image, and a fixed list of 50 predicates (aligning with Visual Genome semantics). For each object pair $(s, o)$, the model is asked to return up to five unidirectional relationships, ranked by plausibility and associated confidence scores. 
These outputs form a multi-relational symbolic prior $\hat{\mathcal{T}} = {(s, p, o)}$, where each object pair $(s, o)$ may be associated with multiple plausible predicates. 
This semantic hypergraph encodes structured hypotheses derived from linguistic and commonsense knowledge. Since the LLM does not guarantee a fixed number of outputs, we normalize the confidence scores per pair to maintain consistent relative weighting. 
\textit{The exact prompt used is provided in the supplementary material. }
This procedure results in a semantically rich hypothesis space with many plausible relationships, including those not in the training set. However, these hallucinated triplets are not grounded in image content—they reflect what \textit{could} be true given object semantics, rather than what \textit{is} true in the scene.

\subsection{Maximization Step: Visual Grounding Model}

To align the hallucinated triplets $\hat{\mathcal{T}}$ with image evidence, we train a visual relationship model $P_\theta(s, p, o \mid I)$ to predict grounded interactions from the image content. We adapt the architecture proposed in IS-GGT~\cite{kundu2023ggt}, due to its ability to combine localized visual-semantic features with global scene context. Concretely, our model is a decoder-only transformer. For each candidate edge $(s, o)$, we construct a query embedding $q_{s,o}$ by concatenating the RoI-pooled visual features of the subject and object, $\phi_v(s)$ and $\phi_v(o)$, with their semantic embeddings $\phi_w(s)$ and $\phi_w(o)$ as $q_{s,o} = [\phi_v(s), \phi_v(o), \phi_w(s), \phi_w(o)]$. 
These query embeddings are passed into the decoder, which attends over a key-value memory derived from frozen DETR~\cite{carion2020end} image features $F_I = \phi_{\text{DETR}}(I)$, providing global scene context. The decoder outputs a predicate distribution $P_\theta(p \mid s, o, I)$ for each query. We supervise the model using only the hallucinated triplets $\hat{\mathcal{T}}$ from the LLM and train it to align symbolic hypotheses with visual content by minimizing:
\begin{equation}
\mathcal{L}_{\text{align}} = -\sum{(s,p,o) \in \hat{\mathcal{T}}} \log P_\theta(p \mid s, o, I).
\end{equation}
This architecture allows the model to resolve ambiguities in the symbolic hypergraph by leveraging both local features of the object pair and global context from the image. During inference, this enables accurate grounding of plausible relationships, even when not observed during training. 
No ground-truth triplets are used at any point in training. Instead, once the model is trained on the hallucinated annotations $\hat{\mathcal{T}}$, we use its predictions to refine the training signal. Specifically, we identify high-confidence triplets predicted by the model—those where the predicate $p$ belongs to the set of seen predicates and the confidence exceeds a fixed threshold $\tau$ (set to 0.8 in our experiments). These newly grounded triplets are accumulated in an auxiliary set $\mathcal{T}_{\text{add}}^{(t)}$ and used to augment the static prior set. We define the full training set at iteration $t$ as $\mathcal{T}_{\text{train}}^{(t)} = \hat{\mathcal{T}} \cup \mathcal{T}_{\text{add}}^{(t)}$. 
The visual grounding model is then fine-tuned on $\mathcal{T}_{\text{train}}^{(t)}$, and the process is repeated, until convergence, i.e., no new triplets are added. 

While the symbolic prior may associate multiple predicates with a single object pair, the visual grounding model predicts one predicate per pair, resolving ambiguity to predict a scene graph.

\begin{table*}[t]
\centering
\resizebox{\textwidth}{!}{
\begin{tabular}{|c|c|c|c|c|c|c|c|c|c|c|}
\toprule
\multirow{2}{*}{\textbf{Approach}} & \multirow{2}{*}{\textbf{Supervision}} 
& \multicolumn{3}{|c|}{\textbf{Seen}} 
& \multicolumn{3}{|c|}{\textbf{Unseen}} 
& \multicolumn{3}{|c|}{\textbf{Mixed}} \\
\cline{3-11}
& & \textbf{mR@20} & \textbf{mR@50} & \textbf{mR@100} & \textbf{mR@20} & \textbf{mR@50} & \textbf{mR@100} & \textbf{mR@20} & \textbf{mR@50} & \textbf{mR@100} \\
\midrule
GGT & Full  & 9.0 & 13.0 & 15.7 & 0.0 & 0.0 & 0.0 & 3.7 & 5.2 & 6.7 \\
FGPL & Full  & 4.1 & 5.8 & 6.7 & 2.0 & 2.4 & 2.4 & 1.9 & 3.2 & 4.3 \\
HiKER-SGG & Full & 7.2 & 9.1 & 10.2 & 0.0 & 0.0 & 0.0 & 3.0 & 4.3 & 5.2 \\
ProtoNet (5-shot) & Full & 8.4 & 10.9 & 12.5 & 3.2 & 6.2 & 6.4 & 2.8 & 5.3 & 6.7 \\
ProtoNet (10-shot) & Full & 7.3 & 9.5 & 11.3 & 9.4 & 10.4 & 10.7 & 3.2 & 5.3 & 7.1 \\
\midrule
GPT4o+ProtoNet (5-shot) & Weak & 8.7 & 11.2 & 12.7 & 8.1 & 11.2 & 11.9 & 5.9 & 9.1 & 11.2 \\
GPT4o+ProtoNet (10-shot) & Weak & 9.5 & 11.4 & 13.2 & \underline{10.3} & 12.4 & 13.5 & 6.5 & 10.3 & 13.1 \\
GPT4o+GGT & Weak & {10.5} & \underline{14.6} & \underline{17.2} & 10.1 & \underline{12.9} & {14.5} & \textbf{8.5} & \underline{11.4} & {14.2} \\
\midrule
GPT-4o (ungrounded) & None & 6.1 & {12.4} & 16.2 & 9.8 & 11.4 & \textbf{16.1} & 4.1 & 8.2 & 12.2 \\
\midrule
\textbf{EM-Grounding (Ours)} & Weak & \underline{11.0} & {14.4} & {16.8} & \textbf{11.7} & \textbf{13.1} & \underline{14.6} & \underline{7.0} & \textbf{11.7} & \underline{15.5} \\
\textbf{EM-Grounding (Ours)} & None & \textbf{12.0} & \textbf{15.9} & \textbf{18.7} & 9.1 & \textbf{13.1} & \underline{14.6} & 6.9 & 11.3 & \textbf{15.8} \\
\bottomrule
\end{tabular}
}
\caption{Predicate classification (PredCls) performance on seen, unseen, and mixed subsets. EM-Grounding consistently outperforms all weakly- and few-shot supervised baselines.}
\label{tab:predcls_main}
\end{table*}

\subsection{Training: Iterative Refinement Loop}

Our refinement strategy is driven by the central hypothesis that reinforcing grounded relationships involving seen predicates improves the model’s ability to generalize to unseen predicates. While the hallucinated triplets $\hat{\mathcal{T}}$ provide broad semantic coverage, they lack visual grounding. By selectively reinforcing high-confidence predictions over known predicates, we provide structurally valid and perceptually supported supervision that helps the model internalize generalizable relational patterns. This distinguishes our approach from generic weak supervision or cross-modal alignment: rather than directly training on noisy pseudo-labels, we iteratively filter and refine grounded structure using a symbolic prior. 
The hallucinated set $\hat{\mathcal{T}}$ remains fixed throughout training. After the visual grounding model $P_\theta$ is trained on this initial supervision, we apply it to all training images to predict new candidate triplets. We retain those that (i) involve predicates in the seen set $\mathcal{P}_{\text{seen}}$, (ii) exceed a confidence threshold $\tau$ (set to 0.8), and (iii) are not already present in $\hat{\mathcal{T}}$ or the cumulative set of previously added grounded triplets. The filtered set is added to an auxiliary pool:

\begin{equation}
\mathcal{T}_{\text{add}}^{(t)} = \left\{(s, p, o) \in \hat{\mathcal{T}}^{(t)} \mid p \in \mathcal{P}_{\text{seen}},\; \text{conf}(s, p, o) > \tau,\; (s, p, o) \notin \hat{\mathcal{T}} \cup \mathcal{T}_{\text{add}}^{(t-1)} \right\}
\end{equation} 
We define the training set at iteration $t$ as \( \mathcal{T}_{\text{train}}^{(t)} = \hat{\mathcal{T}} \cup \mathcal{T}_{\text{add}}^{(t)} \), and fine-tune the model on this combined supervision. The refinement process continues until no new triplets are added. In practice, we observe convergence within 3 iterations. 
Empirically, we find that lowering the confidence threshold to add more triplets degrades generalization to unseen predicates.

\textbf{Implementation Details.} All our experiments utilize GPT-4o as the core Large Language Model. For GGT~\cite{kundu2023ggt} experiments, we use the original paper's pipeline, focusing our training efforts exclusively on the relationship classifier (50 epochs) and the edge decoder (20 epochs). The procedure in IS-GGT is followed for node predictions, as our work centers on open-world predicate generalization rather than object recognition. 
The edge decoder and relation predictor are trained using Adam~\cite{kingma2014adam} with a learning rate of 1e-3 and weight decay of 1e-5. 

All models were trained on a NVIDIA RTX 3090. The supplementary contains detailed implementation information and code.

\section{Experimental Setup}

\textbf{Data.} To evaluate the role of symbolic priors in generalizing beyond annotated relationships, we construct a new benchmark split from the Visual Genome (VG) dataset~\cite{krishna2017visual} tailored for open-world visual relationship detection. 
The goal is to simulate a constrained supervision setting where only a subset of predicates is available during training, while testing generalization to novel relational structures. From the original VG training split, we sample 475 images containing the 29 most frequent predicates, constituting the \textit{seen} predicate set. This results in a lightweight training set with 2,226 annotated triplets—chosen to reflect realistic low-resource conditions and evaluate the effectiveness of symbolic priors rather than distributional co-occurrence. Importantly, the training set is disjoint from any \textit{unseen} predicates. 
For evaluation, we merge the original VG validation and test splits to create a combined set of 5,777 images, comprising 40,884 annotated triplets across 50 predicates. This set is stratified into three mutually exclusive subsets: a \textit{seen-only} split (4,461 images, 29 predicates, 28,322 triplets), an \textit{unseen-only} split (167 images, 19 predicates, 361 triplets), and a \textit{mixed} split (1,149 images, 12,201 triplets) containing at least one seen and one unseen predicate per image. 
This setup allows us to assess performance on: (i) generalization to novel predicates, (ii) compositional reasoning in mixed scenes, and (iii) standard in-distribution predicate prediction. 

Two predicates, ``\textit{says}'' and ``\textit{flying in}'', appear only in the mixed subset as they never occur in isolation.

\textbf{Baselines.} We compare our framework against various baselines spanning different supervision regimes. Supervised baselines include IS-GGT\cite{kundu2023ggt}, FGPL~\cite{lyu2022fine}, and HiKER-SGG~\cite{zhang2024hiker}.

To evaluate the utility of language priors \textit{without} visual grounding, we include an LLM-only baseline, GPT-4o~\cite{hurst2024gpt}, where the model is prompted with object labels to hallucinate triplets directly.

We implement prototypical networks~\cite{snell2017prototypical} (ProtoNet) as few-shot baselines, trained with 10 examples per predicate and experimented with 5/10 shots during inference. 

Finally, we evaluate two weakly supervised baselines: GPT4o+GGT, where GGT is trained directly on LLM-generated triplets \textit{without iterative refinement}, and our model (Grounded-EM), which uses a GGT-style architecture trained with the proposed EM-style iterative grounding. 
All baselines, except FGPL and HiKER-SGG, use the graph sampling and background modeling strategy from GGT~\cite{kundu2023ggt}, which implicitly models the background (no edge) class. This design separates edge selection from predicate classification, allowing models without visual supervision to be fairly evaluated without penalty. 
Complete dataset information and training details for all baselines are in the supplementary.

\textbf{Tasks and Metrics.} We primarily focus on the PredCls and SGCls tasks, as defined in prior works~\cite{zellers2018neural,kundu2023ggt}. PredCls aims to predict the correct relationships between each object pair, given groundtruth bounding boxes and object labels. In SGCls, only groundtruth bounding boxes are provided.

We use mean Recall@K (mR@K) with $K \in {20, 50, 100}$ as our primary evaluation metric, following standard practice~\cite{kundu2023ggt,zhang2024hiker,zhang2022fine}, under the ``graph constraint'' setting. Unlike Recall@K, which is dominated by frequent predicates due to long-tailed annotation distributions, mR@K averages recall across all predicates, making it a better measure of generalization, particularly for rare or unseen predicates. 
mR@K is more informative since we evaluate models on splits defined by seen and unseen \textit{predicates}, rather than triplets ($(s, p, o)$).

\begin{table*}[t]
\centering
\resizebox{\textwidth}{!}{
\begin{tabular}{|c|c|c|c|c|c|c|c|c|c|c|}
\toprule
\multirow{2}{*}{\textbf{Approach}} & \multirow{2}{*}{\textbf{Supervision}} 
& \multicolumn{3}{|c|}{\textbf{Seen}} 
& \multicolumn{3}{|c|}{\textbf{Unseen}} 
& \multicolumn{3}{|c|}{\textbf{Mixed}} \\
\cline{3-11}
& & \textbf{mR@20} & \textbf{mR@50} & \textbf{mR@100} & \textbf{mR@20} & \textbf{mR@50} & \textbf{mR@100} & \textbf{mR@20} & \textbf{mR@50} & \textbf{mR@100} \\
\midrule
GGT & Full  & 5.8 & 7.7 & 9.4 & 0.0 & 0.0 & 0.0 & 2.2 & 3.5 & 4.6\\
FGPL & Full  & 0.2 & 0.2 & 0.3 & 0.0 & 0.0 & 0.0 & 0.0 & 0.0 & 0.0\\
HiKER-SGG & Full & 0.2 & 1.9 & 2.0 & 0.0 & 0.0 & 0.0 & 0.5 & 0.6 & 0.6\\
ProtoNet (5-shot) & Full & 5.5 & 6.5 & 7.5 & 3.6 & 4.3 & 4.3 & 2.4 & 4.0 & 4.9\\
ProtoNet (10-shot) & Full & 4.7 & 6.0 & 6.9 & 5.0 & 5.0 & 5.0 & 2.7 & 4.2 & 5.2\\
\midrule
GPT4o+ProtoNet (5-shot) & Weak & 5.8 & 7.2 & 8.0 & 5.8 & 7.2 & {8.6} & 3.7 & 4.8 & 5.9\\
GPT4o+ProtoNet (10-shot) & Weak & 5.9 & 7.2 & 8.0 & 4.9 & 6.1 & 6.9 & \underline{4.2} & 5.8 & 7.0\\
GPT4o+GGT & Weak & \textbf{6.5} & \textbf{9.1} & \textbf{10.9} & \underline{7.7} & \textbf{8.6} & \textbf{9.5} & 4.0 & \underline{6.1} & 7.8\\
\midrule
GPT-4o (ungrounded) & None & 4.9 & 7.8 & 9.6 & \textbf{7.9} & \underline{8.2} & \underline{8.8} & 3.4 & 5.6 & 8.0\\
\midrule
\textbf{EM-Grounding (Ours)} & Weak & \underline{6.3} & 8.5 & {10.0} & 4.3 & 5.1 & 6.5 & \textbf{4.4} & \textbf{6.9} & \underline{8.4}\\
\textbf{EM-Grounding (Ours)} & None & {6.2} & \underline{8.6} & \underline{10.2} & 4.3 & 5.1 & 6.5 & \textbf{4.4} & \textbf{6.9} & \textbf{8.7}\\
\bottomrule
\end{tabular}
}
\caption{Scene graph classification (SGCls) performance on seen, unseen, and mixed subsets. EM-Grounding consistently outperforms all weakly- and few-shot supervised baselines.}
\label{tab:sgcls_main}
\end{table*}

\section{Results and Analysis}\label{sec:results}

\textbf{Seen predicate classification}. 
Table~\ref{tab:predcls_main} summarizes the results on the PredCls task. 
Despite lacking ground-truth predicate labels during training, both Ours and GPT+GGT achieve competitive or superior performance to fully supervised models on seen predicates. This challenges the conventional assumption that direct supervision is essential for strong relational prediction. Our framework outperforms the supervised GGT across most metrics (mR@20: 12.0 vs. 9.0), highlighting the limitations of supervision under sparse annotation regimes. 
Rather than memorizing biased co-occurrence patterns, our method learns to align symbolic hypotheses with visual signals during training, resulting in more robust representations. 
ProtoNets trained on hallucinated LLM labels also outperform GT-based ProtoNets, suggesting that LLM-derived predicates offer broader coverage.

\textbf{Generalization to unseen predicates.} 
The results on unseen predicate performance underscore the central claim of this work: visual grounding of symbolic priors enables robust generalization to novel relationships. While traditional supervised models like IS-GGT and FGPL, as well as few-shot variants with 5 or 10 annotated samples, perform competitively on seen predicates, they fail to transfer this performance to unseen predicates, often collapsing to near-zero accuracy. GPT-4o provides a strong prior through LLM hallucination, but without grounding, it struggles to resolve visual ambiguities or context-sensitive relationships. 
Our method outperforms all baselines despite not using ground truth annotations by refining LLM-generated hypergraphs through visual feedback.

\textbf{Generalized Prediction.} The mixed split—where both seen and unseen predicates co-occur within the same scene—offers the most realistic and challenging setting, requiring compositional generalization under ambiguity. In this regime, supervised baselines (IS-GGT, FGPL, HiKER-SGG) and few-shot variants trained on only seen predicates exhibit a pronounced failure mode: they tend to overpredict seen relationships while completely ignoring or misclassifying unseen ones, leading to inflated confidence in incorrect edges and substantial drops in recall. Even when given 10 examples per unseen predicate, few-shot methods struggle to integrate novel concepts alongside familiar ones. LLM-only approaches, such as GPT-4o, perform slightly better by generating semantically plausible relationships, but lack the visual grounding necessary to disambiguate contextually relevant edges from distractors. In contrast, our proposed framework (EM-Grounding) significantly outperforms all baselines, demonstrating the ability to predict both seen and unseen predicates correctly.

\begin{figure*}
    \begin{tabular}{ccc}
        \includegraphics[width=0.3\textwidth]{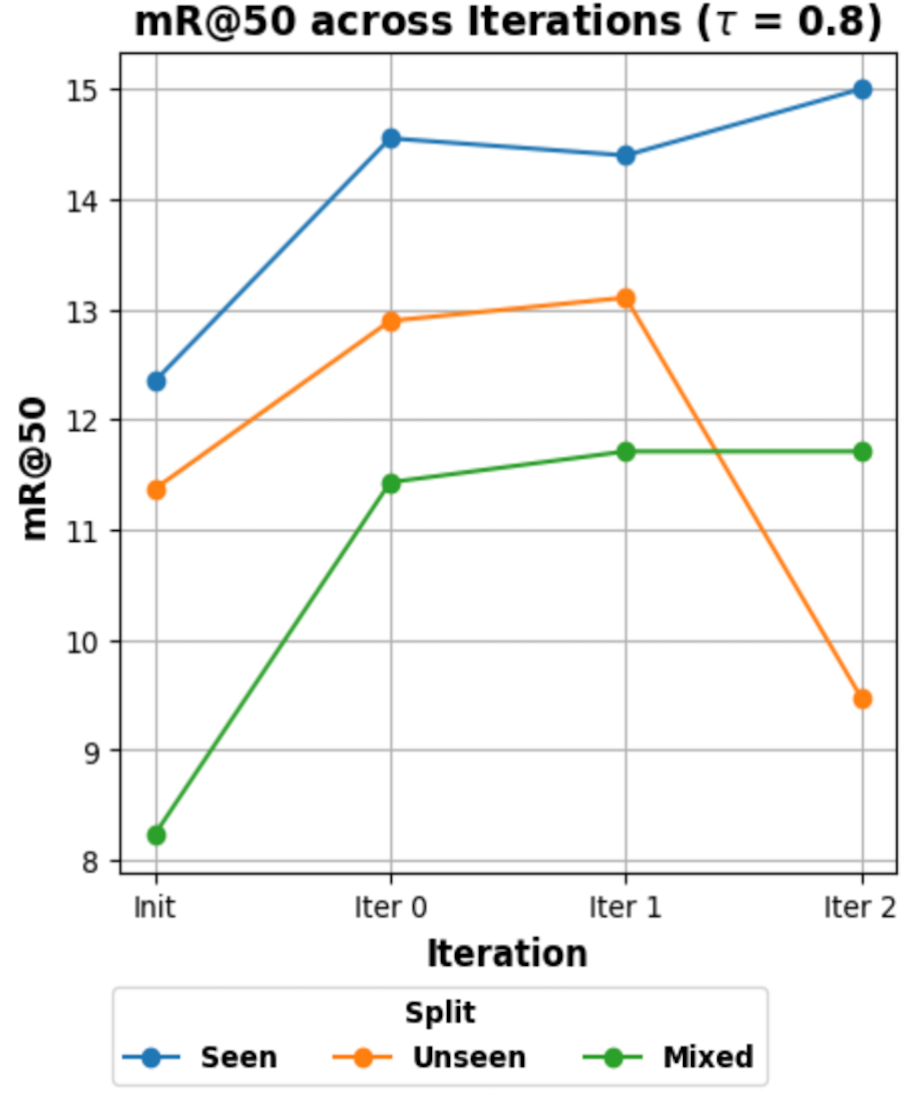} &  
        \includegraphics[width=0.3\textwidth]{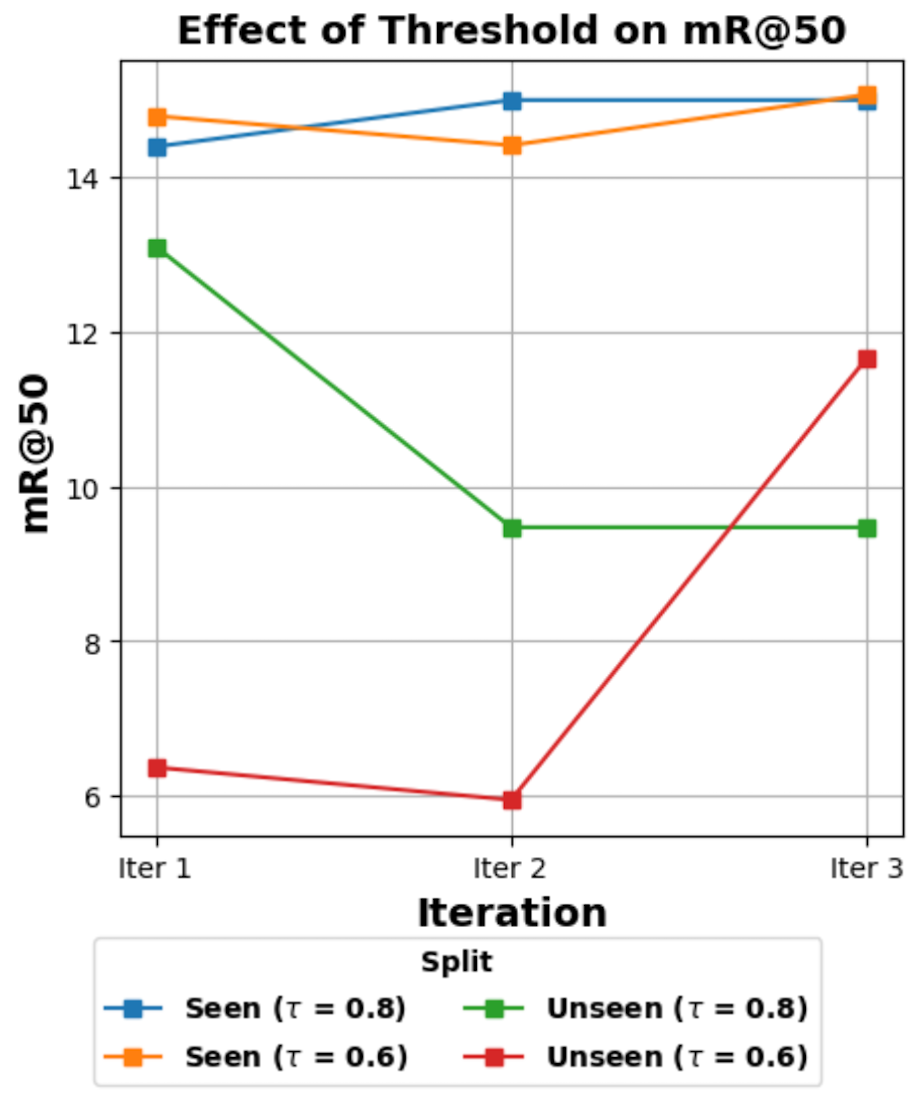} & 
        \includegraphics[width=0.3\textwidth]{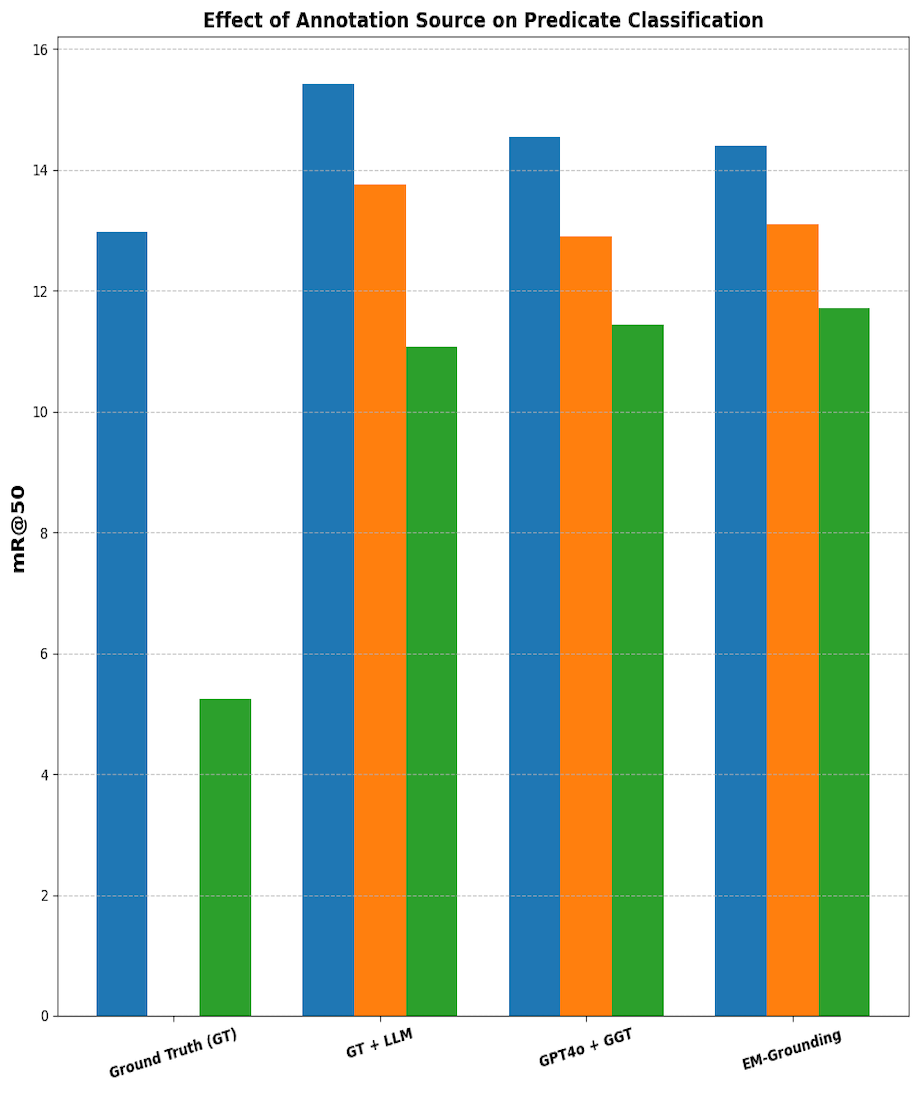} \\
        (a) & (b) & (c)\\
    \end{tabular}
    \caption{\textbf{Ablation studies} showing the effect of (a) iterative refinement, (b) threshold ($\tau$) settings, and (c) annotation sources on mR@50 for seen, unseen, and mixed predicate recognition. 
    }
    \label{fig:ablations}
\end{figure*}

\textbf{Bridging the Supervision Gap with Symbolic Priors.} To contextualize our performance, we evaluate EM-Grounding on the official Visual Genome test set, despite being trained on just 475 images covering 29 seen predicates with only 2.2k annotated triplets. In contrast, prior state-of-the-art models are trained on over 57k images with 405k triplets spanning all 50 predicates. Despite this 100× supervision gap, our model achieves mR@50 of 11.8 and mR@100 of 17.2, outperforming fully supervised baselines like IMP+ (9.8 / 10.5) and Neural Motifs (14.0 / 15.3), and approaching VCTree (17.9 / 19.4). While debiasing-based approaches like HiKeR-SGG (39.3 / 41.2), PCPL (35.2 / 27.8), and GBNet (19.3 / 20.9) achieve higher scores with access to the full dataset and specialized mitigation strategies.

Full comparisons are included in the supplementary material.

\begin{figure*}
\begin{tabular}{c}
    \includegraphics[width=0.95\textwidth]{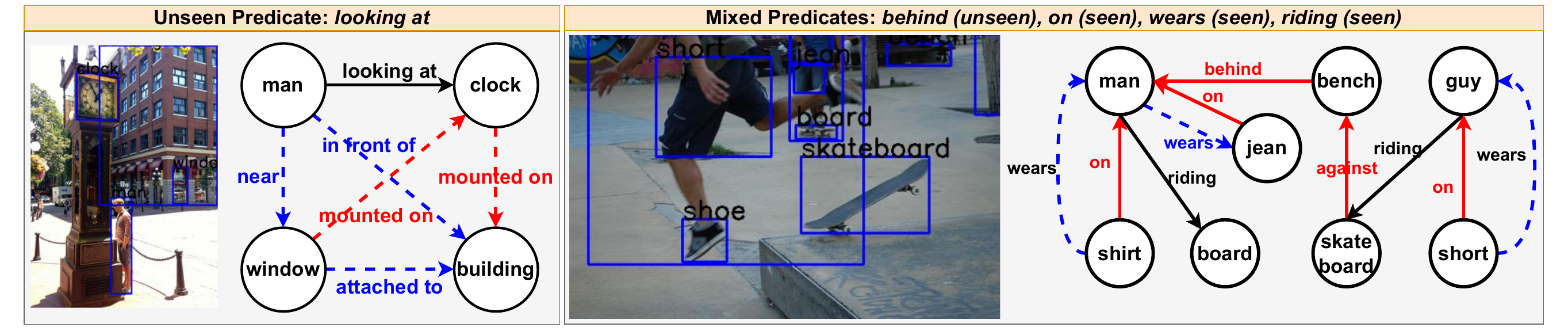}\\
    \includegraphics[width=0.95\textwidth]{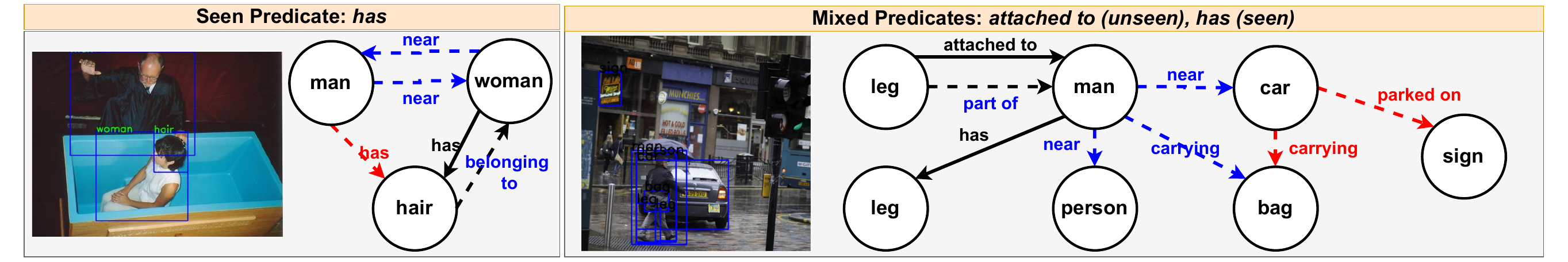}\\
\end{tabular}
\label{fig:qual}
\caption{\textbf{Qualitative results} on different settings: unseen predicates only (top left), seen only (bottom left), and mixed (top right and bottom right). Correctly predicted groundtruth (GT) edges are in solid black, missed GT edges in solid red, visually correct but unannotated edges in dashed blue, and visually incorrect, unannotated edges in dashed red. }
\end{figure*}

\textbf{Scene Graph Classification Performance.}
We report the performance of all baselines in the SGCls task in Table~\ref{tab:sgcls_main}. 
In the SGCls setting, where both object labels and predicates must be predicted, our EM-Grounding framework achieves the strongest performance on the mixed split (mR@100 = 8.7), highlighting its ability to reconcile seen and unseen relationships within a single graph. While GPT4o+GGT achieves slightly higher scores on the isolated seen and unseen subsets (e.g., mR@100 = 10.9 and 9.5, respectively), EM-Grounding excels when both distributions are present, which better reflects real-world inference. 

Compared to ProtoNet variants, 
EM-Grounding offers stronger overall generalization, even without relying on predicate-level supervision. Although GPT-4o retrieves many plausible edges (e.g., mR@20 = 7.9 for unseen), its precision deteriorates at higher recall thresholds.

\textbf{Impact of Interaction Modeling.} To evaluate the importance of interaction priors in EM-Grounding, we remove the GGT-trained interaction predictor and use a fully connected scene graph during training and inference instead. This setup eliminates the need to model edge presence, a component typically learned from visual cues, making it more aligned with unsupervised regimes where such information is unavailable. As shown in Tables~\ref{tab:predcls_main} and~\ref{tab:sgcls_main}, this no-interaction variant still performs competitively, achieving mR@50 of {13.1} on unseen predicates in PredCls and {5.1} in SGCls. 

\textbf{Impact of Refining Iterations.} Figure~\ref{fig:ablations}(a) shows the impact of refining iterations on the performance, beginning with LLM-generated graphs (init).  Iterative hallucinate-and-ground refinement improves performance, particularly on the mixed set, which benefits from progressively better coverage of both seen and unseen predicates. Gains saturate after two iterations, indicating diminishing returns.

\textbf{Impact of Threshold ($\tau$).} As can be seen from Figure~\ref{fig:ablations}(b), a stricter threshold ($\tau=0.8$) yields more reliable triplets, resulting in better unseen and mixed performance compared to $\tau=0.6$, which admits lower-quality annotations that degrade generalization, especially to rare predicates.

\textbf{Impact of LLM-based labeling.} Figure~\ref{fig:ablations}(c) shows that while combining GT with LLM-hallucinated annotations boosts seen performance, EM-Grounding trained without any GT still outperforms all other weak and few-shot variants, demonstrating the strength of symbolic priors in low-label regimes.

\textbf{Qualitative Analysis}. Figure~\ref{fig:qual} provides qualitative visualizations illustrating that EM-Grounding accurately recovers annotated and unannotated but visually valid relationships across seen, unseen, and mixed predicate settings. In unseen cases, the model correctly grounds novel interactions (e.g., “looking at”) without prior supervision. Mixed-predicate examples highlight its ability to reconcile familiar and novel relations within the same scene. The model often predicts visually correct relationships not in the groundtruth, highlighting the utility of grounding LLM-driven priors.

\section{Discussion, Limitations, and Future Work}
We present EM-Grounding, a novel framework for generalized visual relationship detection that leverages symbolic priors from LLMs and grounds them through iterative EM-style refinement. By treating hallucinated triplets as a structured hypothesis space, we selectively align them with visual evidence, enabling generalization to unseen predicates with limited supervision. EM-Grounding outperforms all weakly- and few-shot baselines across tasks. 
While EM-Grounding offers a scalable path toward generalized scene understanding, one must mitigate inherited biases from LLMs.

\textbf{Limitations.} Despite its effectiveness, EM-Grounding has some limitations. First, it assumes access to accurate object detections; errors at this stage can misguide priors and degrade predictions. 
Second, the symbolic prior is derived solely from object labels and lacks visual and spatial context, which can lead to implausible or overly generic triplets. 
Finally, our current focus is generalized predicate learning, which is restricted to a pre-defined label space. True open-world learning, i.e., scenarios involving unseen predicates \textit{and} objects, is not yet evaluated. 

\textbf{Future Work.} Addressing these limitations opens several paths forward. Incorporating spatial cues into LLM prompts can yield more grounded priors. 
Additionally, structured uncertainty quantification in the refinement loop could help manage confidence vs. coverage. Expanding to open-vocabulary benchmarks will further test generalization. We aim to adapt EM-Grounding to these open-world tasks and provide a scalable and extensible basis for symbolically grounded visual understanding. 

\section{Acknowledgements}
This research was supported in part by the US National Science Foundation grants IIS 2348689 and IIS 2348690.

\bibliographystyle{abbrvnat}
\bibliography{neurips_2025}

\clearpage

\maketitle

\section{Appendix A}
This supplementary material provides additional details about the dataset and implementation of various baselines. We also provide additional results for scene graph detection, visualizations and analysis of per class performance to further support our approach. Additionally we also supplement our qualitative results with more images across all the three tasks and splits. Moreover, we share the code as well as the image ids for all our splits in the attached zip file.

\section{Dataset}

Scene graph generation has been researched extensively over the years and quite a few benchmark datasets have been proposed. Although the traditional train, val and test splits are a good starting point to analyze the model's performance, they are not sufficient to measure the impact of the model in the wild where there are "unknowns" and "unseens". To deal with this, especially in the context of predicates, seen and unseen splits have been proposed in recent times where the top-K predicates go under the seen list and the rest of them go in the unseen split. The images are then split according to these criteria.

\begin{figure}[t]
\begin{floatrow}

\capbtabbox{%
  \resizebox{0.3\textwidth}{!}{%
    \begin{tabular}{@{}lccc@{}}
    \toprule
    \textbf{Split} & \textbf{\#Images} & \textbf{\#Predicates} & \textbf{\#Triplets} \\
    \midrule
    Train        & 475   & 29  & 2,226 \\
    Val - Seen   & 4,461 & 29  & 28,322 \\
    Val - Unseen & 167   & 19  & 361 \\
    Val - Mixed  & 1,149 & 50  & 12,201 \\
    \bottomrule
    \end{tabular}
  }
}{%
  \caption{Dataset statistics across evaluation subsets.}
  \label{tab:statistics}
}

\ffigbox{%
  \includegraphics[width=0.65\textwidth]{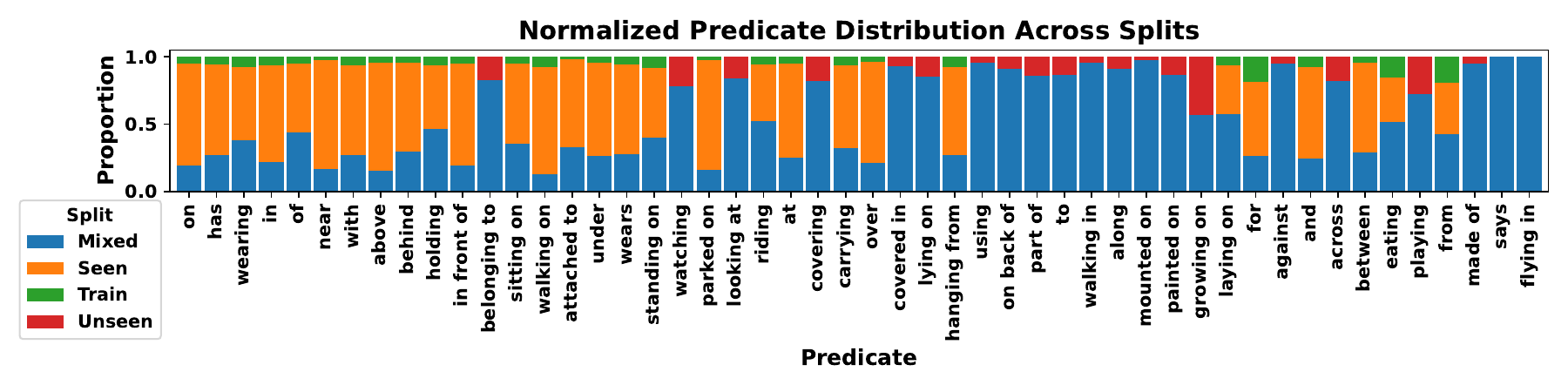}
}{%
  \caption{Normalized predicate distribution across subsets.}
  \label{fig:histogram}
}

\end{floatrow}
\end{figure}

Even though this split makes sense, we believe there is still room for further segregation of the images so that more detailed performance metrics can be extracted. We take an intuitive approach and make three splits instead of two: seen, unseen and mixed. Here seen and unseen refers to seen-only and unseen-only, which means all the images in the unseen split have only unseen predicates. Similarly, the images in seen split don't have any unseen predicates whereas the mixed split contains images which have at least one seen and one unseen predicate. This setup can be more effective for analyzing the impact of the existence of seen predicates on unseen predicates in any given scene. As described in the main paper we designate a subset of 475 images from the Visual Genome train set as our train set here. The 29 predicates here belong to the seen set and the rest of them belong to the unseen set. Our unseen-only split has 19 predicates since the rest of the two predicates always co-occur with seen predicates. The mixed split contains all 50 predicates. In Figure ~\ref{fig:histogram} we show the detailed statistics of our dataset. The image ids for train, seen, unseen and mixed splits have been shared in the zip file.

\newfloatcommand{capbtabbox}{table}[][0.3\textwidth]

\section{Baselines}
Below we provide additional details about the implementation of various baselines used in our paper.

\subsection{GGT Details}
IS-GGT proposed a more efficient, generative transformer-based approach to Scene Graph Generation. By using one transformer to first sample the most probable relationships (edges) and another to classify the predicates on only those sampled edges, the method reduces the computational overhead associated with classifying every possible inter-object relationship. This generative sampling step allows for more efficient inference compared to traditional exhaustive classification methods, while still achieving competitive performance on the Visual Genome dataset, even outperforming some state-of-the-art methods in mean recall. This decoupling between edge sampling and predicate classification is one of the major reasons for choosing GGT as the base model in our own approach. This allows for fair evaluation of the predicate classifier and see how EM based grounding impacts it. 

For both our approach and for the GGT baseline, we directly borrow the pipeline from the original paper for training and evaluation. We keep the architecture and hyperparmeters same as in the paper. Only the graph sampling decoder and the predicate classifier are trained from scratch. The different set of edges obtained from this trained graph sampler under PredCls, SGCls and SGDet settings are used for the inference of all the baselines except FGPL and HiKER-SGG. We share the code in the attached zip file. 

\subsection{LLM Details}
We use GPT4o as the LLM in all our experiments. The below prompt was used for generating multi-relational fully-connected scene graphs for all the splits:

\newtcblisting{PromptBox}{
  colback=gray!5,
  colframe=black,
  listing only,
  breakable,
  enhanced,
  fonttitle=\bfseries,
  title=Prompt Instructions,
  listing options={
    basicstyle=\ttfamily\footnotesize,
    breaklines=true,
    breakindent=0pt,
    columns=fullflexible,
    showstringspaces=false
  }
}

\begin{PromptBox}
Using your prior knowledge of the spatial arrangement of scenes, visualize a realistic scene which has a list of objects that I give you. Now if we pick any two objects from this list, they will have a relationship based on their placement in the scene. So, if I give you a list of objects and a list of pairs from this list of objects, your task is to visualize the scene containing these objects and give me the 5 most likely relationships along with a confidence score for each pair based on that scene.

Note that you can pick the relationships only from the predicate list: ["and", "says", "belonging to", "over", "parked on", "growing on", "standing on", "made of", "attached to", "at", "in", "hanging from", "wears", "in front of", "from", "for", "watching", "lying on", "to", "behind", "flying in", "looking at", "on back of", "holding", "between", "laying on", "riding", "has", "across", "wearing", "walking on", "eating", "above", "part of", "walking in", "sitting on", "under", "covered in", "carrying", "using", "along", "with", "on", "covering", "of", "against", "playing", "near", "painted on", "mounted on"].

Also, one constraint is that the chosen relationship must be unidirectional. If I give you a pair such as 'fruit', 'tree' then you can choose 'growing on' as one of the relationships since fruit can grow on tree. But if I give you 'tree', 'fruit' then you can't choose 'growing on' as one of the relationships since tree can't grow on fruit. So, the order of the pair is important while choosing the relationship.

As an example, list of objects: 'human', 'tree', 'fruit'; list of pairs: ('human','tree') ('fruit', 'tree') ('tree', 'fruit')

For this, your output format should be a simple list like below which will have all the pairs:
1. ('human','tree'); 'under',0.9; 'near',0.9; 'in front of',0.8; 'behind',0.8; 'looking at',0.6  
2. ('fruit','tree'); 'growing on',0.9; 'hanging from',0.9; 'attached to',0.9; 'under',0.8; 'near',0.8  
3. ('tree','fruit'); 'over',0.9; 'near',0.9; 'attached to',0.9; 'behind',0.6; 'across',0.5
\end{PromptBox}
\FloatBarrier

In all our experiments we only use the predicate with the highest score from these predictions in order to get the triplets. For fair comparison we use the GGT graph decoder to obtain the edges and then filtering the GPT4o triplets to include only these edges for computing metrics.

\subsection{ProtoNet Details}
ProtoNets have been proven to show very good generalization in few shot setups. So, we used them as our few-shot baseline models. We followed standard protoNet pipeline with euclidean metric for training and testing. We designate 10 shots and 20 queries during training. For final inference, we compute a set of global prototypes from random images in a 5-shot and 10-shot setup. We use these prototypes as anchors for predicate classification. Additionally we also train these models on the GPT4o generated data for comparing with our approach.

For the architecture, we develop a relation embedding model which first processes global image features using a transformer encoder. Concurrently, it fuses semantic and visual features for both the subject and object through dedicated inter-modal (semantic-visual) and intra-modal (semantic-semantic, visual-visual) fusion modules. The resulting object representations then undergo cross-attention, and are finally combined with the global image context to produce the final relation embedding.

The global image features are extracted using DETR. We use faster-RCNN and BERT for obtaining visual and semantic features of the objects respectively. We first use the GGT graph decoder to get the edges before passing them onto our protoNet for further processing. The code is attached in the zip file.

\subsection{FGPL and HiKER-SGG Details}
Fine-Grained Predicates Learning (FGPL)\footnote{\url{https://github.com/XinyuLyu/FGPL}} tackles the challenge of fine-grained predicate ambiguity in scene graphs by introducing a Predicate Lattice and specific discriminating losses. This model-agnostic approach aims to differentiate hard-to-distinguish predicates, significantly boosting mean recall on predicate classification tasks.

Hierarchical Knowledge Enhanced Robust Scene Graph Generation (HiKER-SGG)\footnote{\url{https://github.com/zhangce01/HiKER-SGG}} provides a robust baseline for scene graph generation in corrupted visual environments, utilizing a hierarchical knowledge graph to refine its predictions from coarse to fine-grained levels. This approach shows superior zero-shot performance on corrupted images and strong results on standard SGG tasks.

Due to their superior performance on predicate classification task, we decided to use them as our supervised baselines. For fair comparison we train and evaluate both of them on our proposed splits. We utilize the official code-bases provided by the authors for both these models for training and evaluation, keeping all the parameters same. FGPL is trained and evaluated on all three settings but HiKER-SGG is only trained on PredCls and SGCls modes in adherence to the original paper.

\section{Additional Results and Discussion}
\subsection{Per Class Recall Analysis}
\begin{figure}[htbp]
    \centering
    \begin{tabular}{c}
        \includegraphics[width=0.9\textwidth]{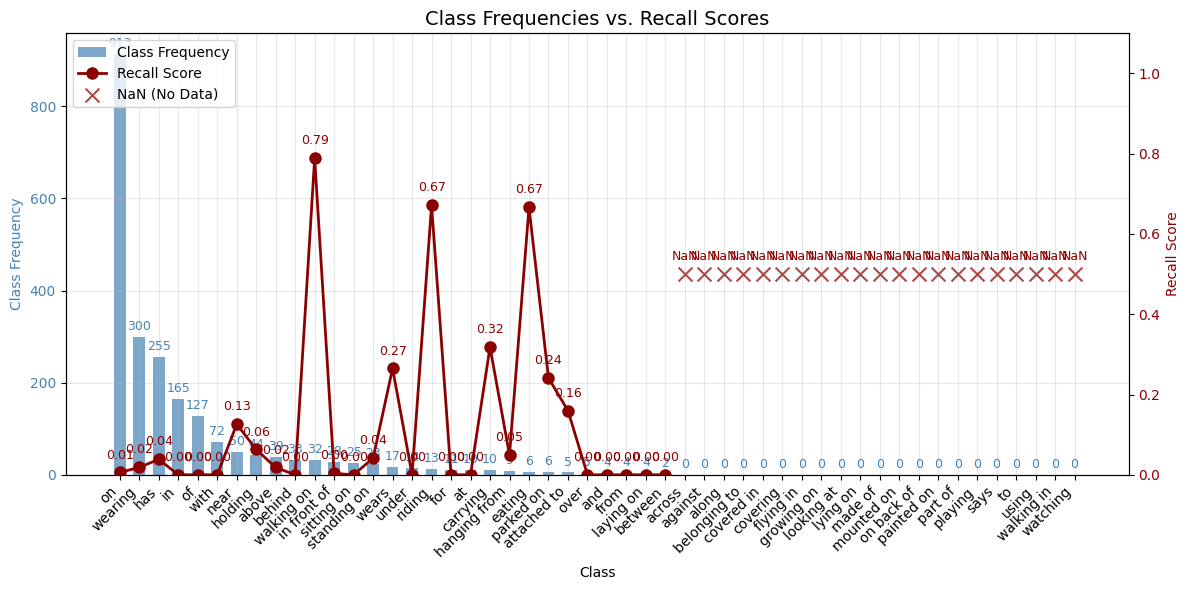} \\
        (a) \\
        \includegraphics[width=0.9\textwidth]{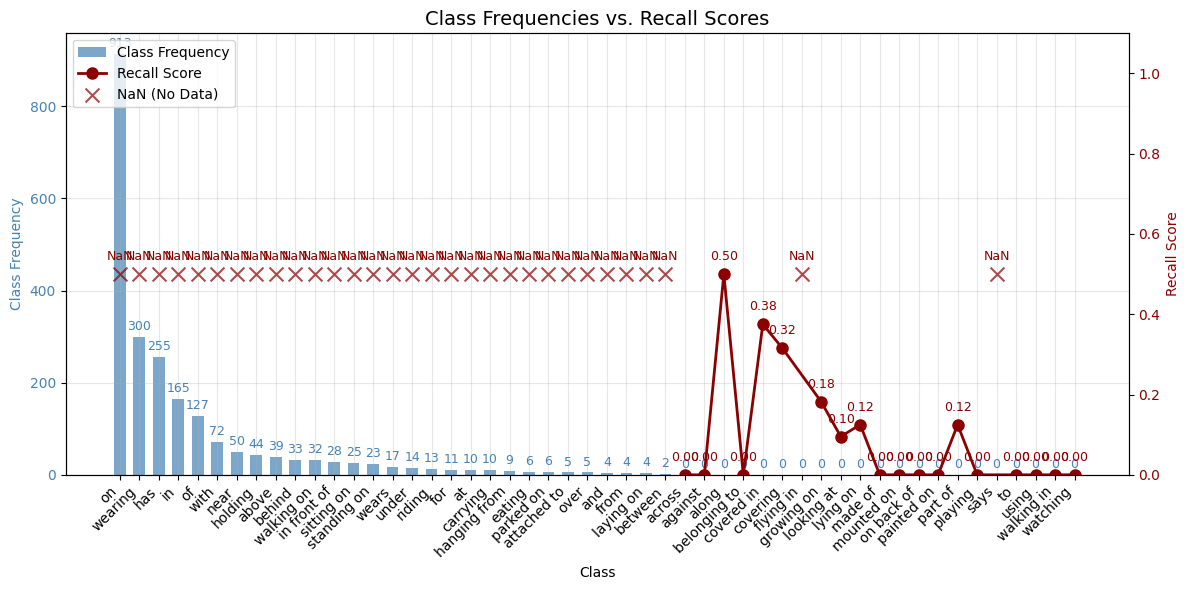} \\
        (b) \\
        \includegraphics[width=0.9\textwidth]{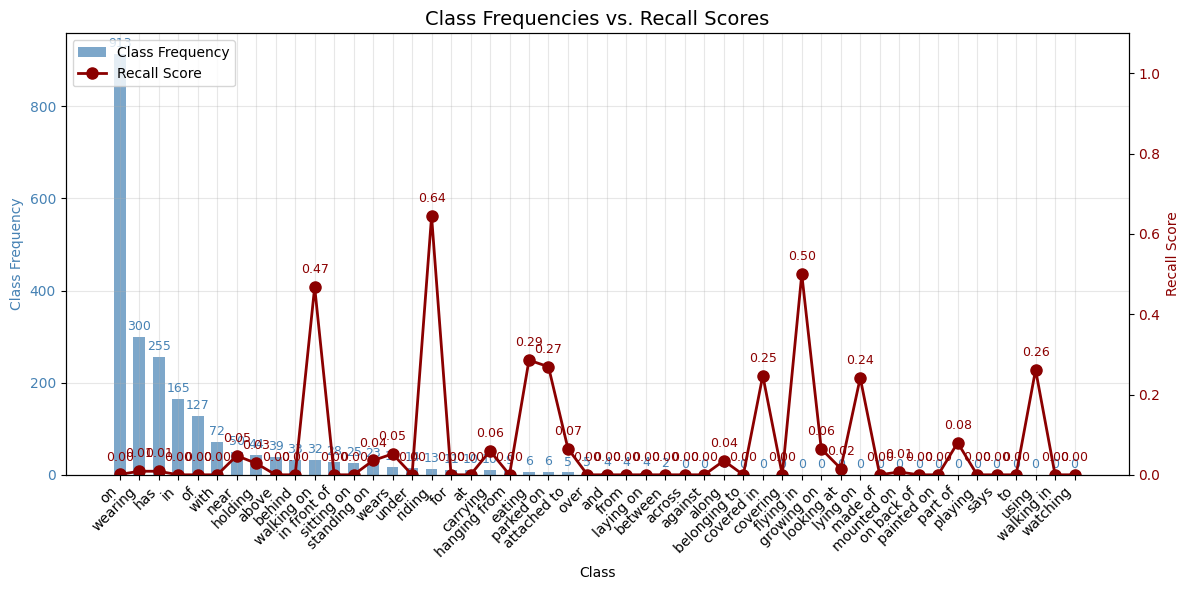} \\
        (c) \\
    \end{tabular}
    \caption{\textbf{Per class recall@20} for (a) seen, (b) unseen, and (c) mixed predicate classification. All the plots also show the histograms of predicate counts in our train set.}
    \label{fig:per_class_recall}
\end{figure}

In Figure ~\ref{fig:per_class_recall}, the per class recalls are shown for each split with histograms of train predicate counts in the background. We can see from these plots that unseen predicates have higher recall scores in general in the mixed split when compared with the unseen split. Whereas the seen predicates scores seem to dip in the mixed split compared to the seen split. Although further analysis into the triplets needs to be performed to confirm this, the results so far show that that the existence of seen predicates in general have a positive impact on the unseen predicates thereby boosting the model performance on them. Regardless, this definitively highlights the value of our proposed `mixed split' for a more insightful evaluation of model generalization and robustness when encountering novel elements in open-world settings.

\subsection{Scene Graph Detection Results}
Even though the main focus of our paper is on PredCls and SGCls tasks, we evaluate our model on Scene Graph Detection task (SGDet) too and provide the metrics in Table~\ref{tab:sgdet_main} for soundness sake. Notably, our approach still consistently outperforms all the baselines even in this setting. Interestingly, ungrounded GPT4o faces the biggest dip in the performance when using this setup where the scores for unseen drop to 0. ProtoNet and GPT4o+GGT still perform better than other fully supervised baselines. These results demonstrate the robustness and efficacy of our model across all three scene graph tasks.

\begin{table*}[t]
\centering
\resizebox{\textwidth}{!}{
\begin{tabular}{|c|c|c|c|c|c|c|c|c|c|c|}
\toprule
\multirow{2}{*}{\textbf{Approach}} & \multirow{2}{*}{\textbf{Supervision}} 
& \multicolumn{3}{|c|}{\textbf{Seen}} 
& \multicolumn{3}{|c|}{\textbf{Unseen}} 
& \multicolumn{3}{|c|}{\textbf{Mixed}} \\
\cline{3-11}
& & \textbf{mR@20} & \textbf{mR@50} & \textbf{mR@100} & \textbf{mR@20} & \textbf{mR@50} & \textbf{mR@100} & \textbf{mR@20} & \textbf{mR@50} & \textbf{mR@100} \\
\midrule
GGT & Full  & 3.1 & 4.9 & 5.8 & 0.0 & 0.0 & 0.0 & 1.5 & 2.4 & 3.2 \\
FGPL & Full  & 0.0 & 0.0 & 0.0 & 0.0 & 0.0 & 0.0 & 0.0 & 0.0 & 0.0 \\
HiKER-SGG & Full & -- & -- & -- & -- & -- & -- & -- & -- & -- \\
ProtoNet (5-shot) & Full & 4.3 & 5.3 & 5.7 & 2.2 & 2.3 & 2.7 & 2.1 & 2.7 & 3.2 \\
ProtoNet (10-shot) & Full & 3.9 & 5.1 & 5.8 & 1.7 & 1.8 & 1.8 & 1.8 & 2.7 & 3.5 \\
\midrule
ProtoNet (5-shot) & Weak & 4.2 & 5.3 & 5.7 & 1.1 & 1.9 & 3.2 & 2.0 & 2.8 & 3.6 \\
ProtoNet (10-shot) & Weak & 4.4 & 5.7 & 6.2 & 2.6 & 4.1 & 5.2 & 2.2 & 3.9 & 4.9 \\
GPT4o+GGT & Weak & 4.7 & 6.0 & \underline{7.1} & \underline{5.7} & \underline{7.5} & 7.5 & \underline{2.8} & \textbf{4.4} & \textbf{5.3} \\
\midrule
GPT-4o (ungrounded) & None & 0.7 & 1.0 & 1.1 & 0.0 & 0.0 & 0.0 & 0.5 & 0.9 & 1.1 \\
\midrule
\textbf{EM-Grounding (Ours)} & Weak & \textbf{5.0} & \underline{6.3} & \underline{7.1} & 5.4 & 6.7 & \underline{7.6} & \textbf{3.0} & \underline{4.1} & \underline{5.2} \\
\textbf{EM-Grounding (Ours)} & None & \underline{4.9} & \textbf{6.4} & \textbf{7.3} & \textbf{6.0} & \textbf{8.0} & \textbf{8.3} & 2.7 & \underline{4.1} & \textbf{5.3} \\
\bottomrule
\end{tabular}
}
\caption{Scene graph detection (SGDet) performance on seen, unseen, and mixed subsets. EM-Grounding consistently outperforms all weakly-supervised and few-shot baselines. 
}
\label{tab:sgdet_main}
\end{table*}

\subsection{Comparision With SOTA}
Furthermore, we evaluate our approach on the entire original test set to compare against other baselines as show in Table ~\ref{tab:sota_comp}. Interestingly, despite being trained on a dramatically smaller dataset (only 475 images and 29 predicates) and under weak supervision, our approach achieves competitive performance across all tasks. Notably, it surpasses early fully-supervised models like IMP+ and Neural Motifs in both PredCls and SGCls, and performs comparably in SGDet. This is particularly impressive considering those baselines were trained on the full 57k-image dataset with complete annotations. Our method even outperforms GGT (Subset), which uses the same training data but under full supervision, demonstrating the effectiveness of our weak supervision strategy. These results highlight the strong generalization and efficiency of our model in low-data, low-supervision regimes.

\begin{table}[t]
\centering
\resizebox{\textwidth}{!}{
\begin{tabular}{|c|c|c|c|c|c|c|c|c|}
\toprule
\multirow{2}{*}{\textbf{Model}} & \multirow{2}{*}{\textbf{Supervision}} & \multirow{2}{*}{\textbf{Train Set}} 
& \multicolumn{2}{|c|}{\textbf{PredCls}} 
& \multicolumn{2}{|c|}{\textbf{SGCls}} 
& \multicolumn{2}{|c|}{\textbf{SGDet}} \\
\cline{4-9}
 & & & \textbf{mR@50} & \textbf{mR@100} & \textbf{mR@50} & \textbf{mR@100} & \textbf{mR@50} & \textbf{mR@100} \\
\midrule
IMP+                & Full (GT)     & 57k images / 50 predicates     & 9.8  & 10.5  & 5.8  & 6.0  & 3.8  & 4.8  \\
Neural Motifs       & Full (GT)     & 57k images / 50 predicates     & 14.0 & 15.3  & 7.7  & 8.2  & 5.7  & 6.6  \\
VCTree              & Full (GT)     & 57k images / 50 predicates     & 17.9 & 19.4  & 10.1 & 10.8 & 6.9  & 8.0  \\
PCPL                & Full (GT)     & 57k images / 50 predicates     & 35.2 & 37.8  & 18.6 & 19.6 & 9.5  & 11.7 \\
G2S-Transformer     & Full (GT)     & 57k images / 50 predicates     & 31.9 & 34.2  & 18.5 & 19.4 & 14.8 & 17.1 \\
\textbf{GGT (Full)}          & Full (GT)     & 57k images / 50 predicates     & 26.4 & 31.9  & 15.8 & 18.9 & 9.1  & 11.3 \\
\textbf{GGT (Subset)}        & Full (GT)     & 475 images / 29 predicates     & 6.0  & 7.6   & 4.0   & 5.0   & 2.5   & 3.1   \\
\textbf{Ours (Weak)}         & Weak (GPT)    & \textbf{475 images / 29 predicates}     & \textbf{11.7} & \textbf{14.7}  & \textbf{7.0}  & \textbf{8.5}  & \textbf{4.3}  & \textbf{5.3}  \\
\bottomrule
\end{tabular}
}
\caption{Scene graph generation performance (mean Recall @50 and @100) for Predicate Classification (PredCls), Scene Graph Classification (SGCls), and Scene Graph Detection (SGDet) under different supervision settings and training set sizes. Metrics for our approach have been represented by boldface.}
\label{tab:sota_comp}
\end{table}

\subsection{Additional Qualitative Analysis}
\begin{figure*}
    \centering
    \begin{tabular}{cccc}
        \includegraphics[width=0.24\textwidth]{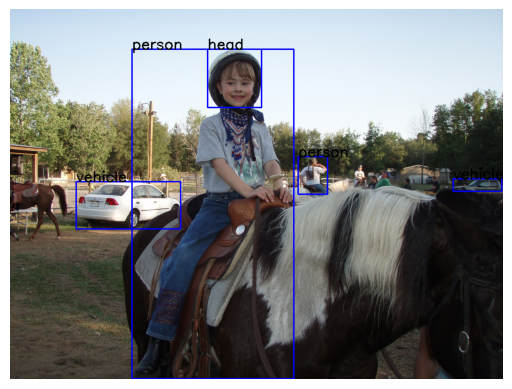} &  
        \includegraphics[height=2.8cm]{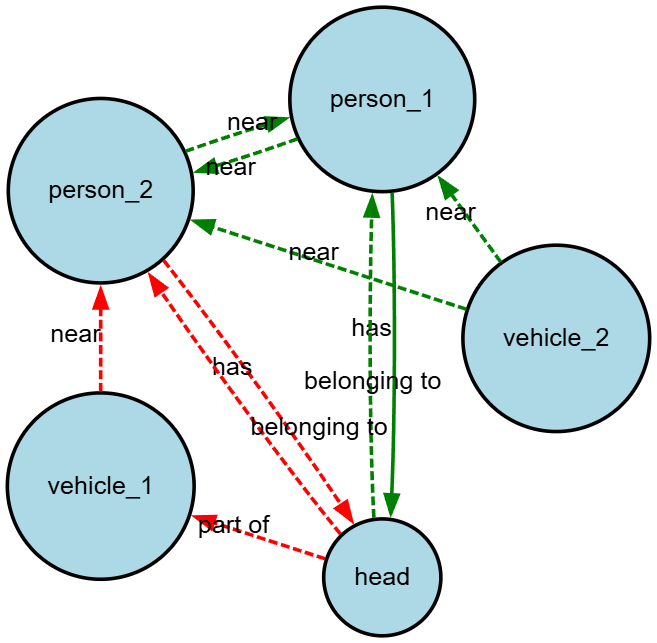} & 
        \includegraphics[width=0.24\textwidth]{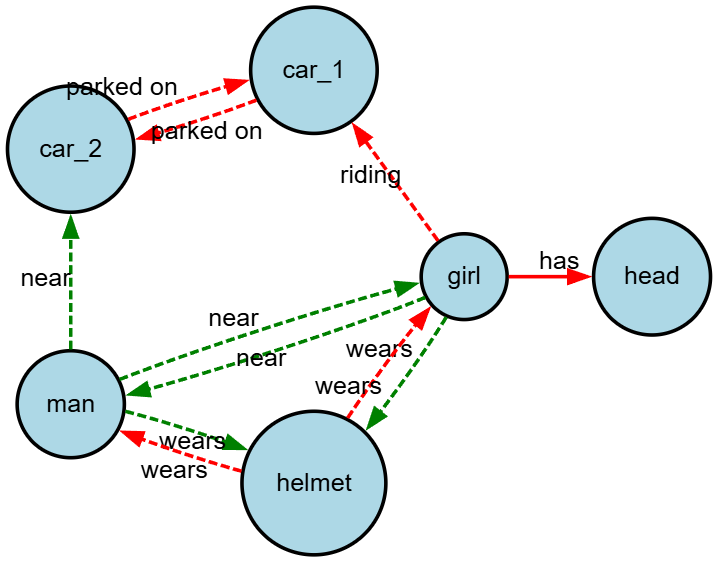} &
        \includegraphics[height=2.8cm]{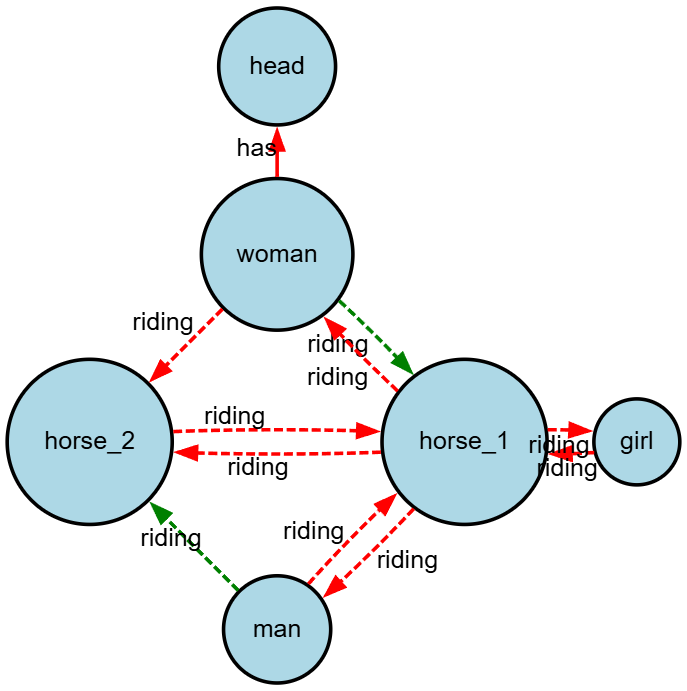} \\
        (a) & (b) & (c) & (d) \\[0.8em]
        
        \includegraphics[width=0.24\textwidth]{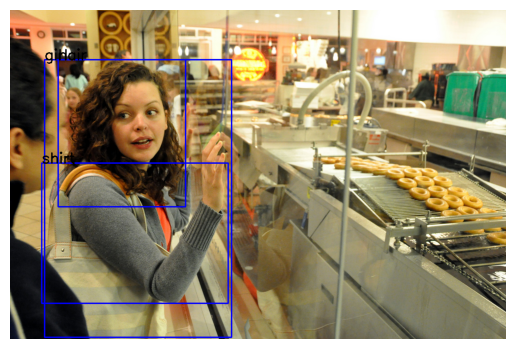} &  
        \includegraphics[width=0.24\textwidth]{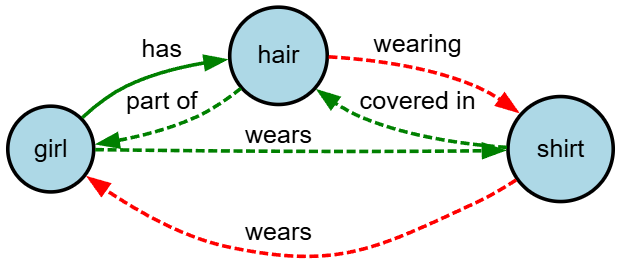} & 
        \includegraphics[width=0.24\textwidth]{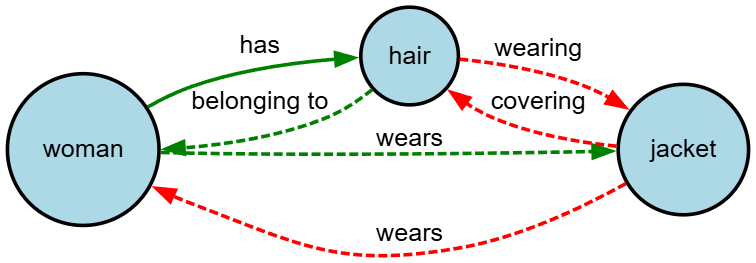} &
        \includegraphics[width=0.24\textwidth]{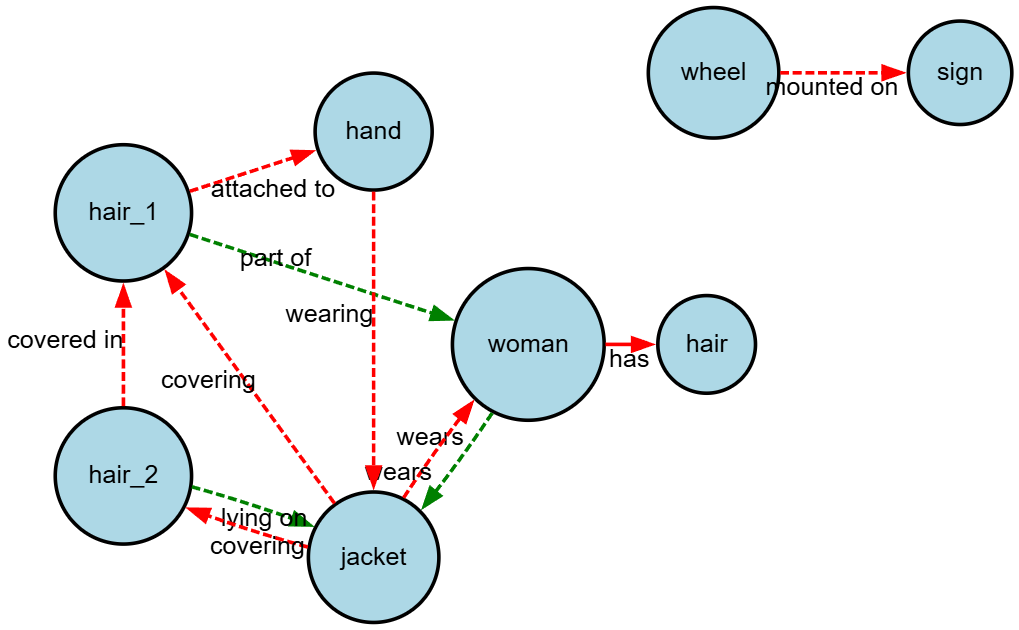} \\
        (e) & (f) & (g) & (h) \\[0.8em]
        
        \includegraphics[height=3cm]{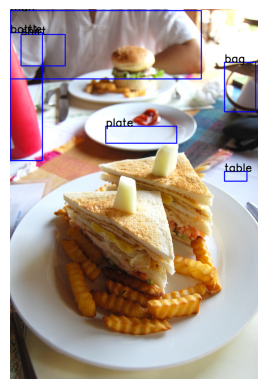} &  
        \includegraphics[width=0.24\textwidth]{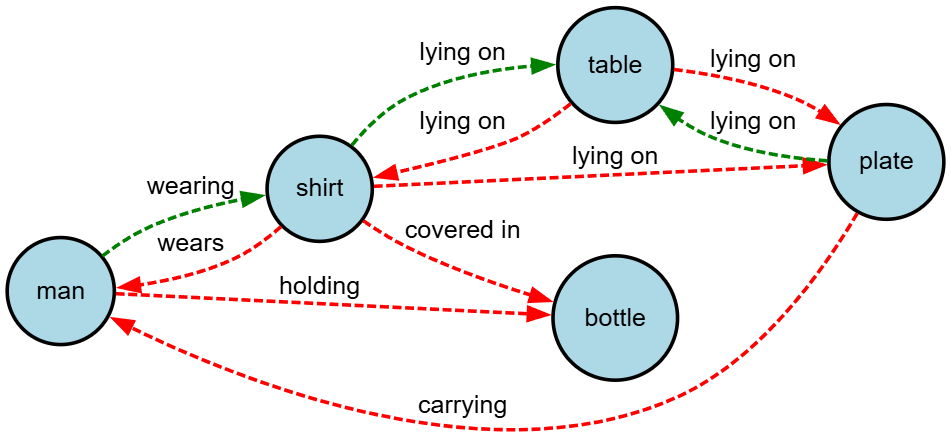} & 
        \includegraphics[width=0.24\textwidth]{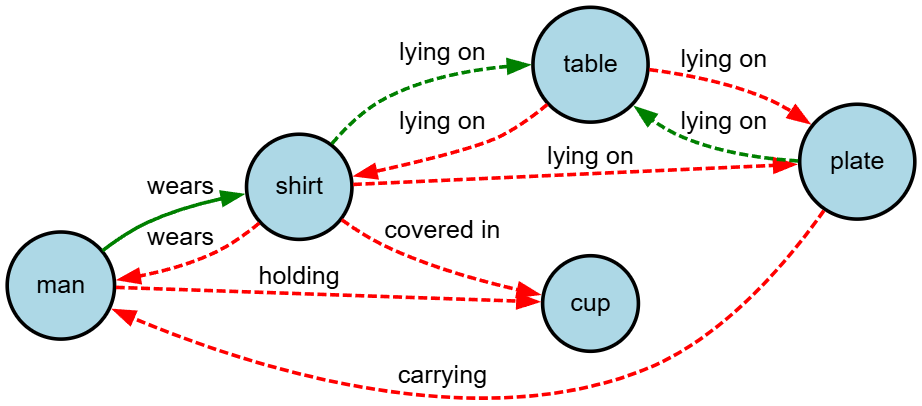} &
        \includegraphics[width=0.24\textwidth]{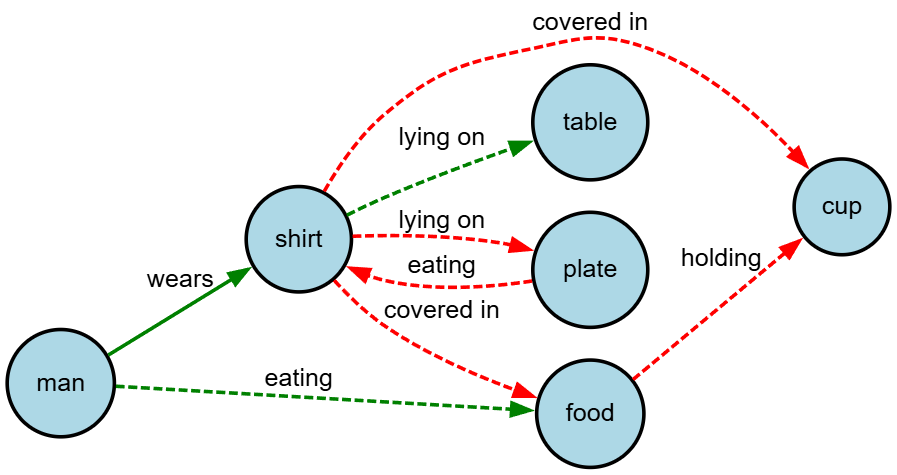} \\
        (i) & (j) & (k) & (l) \\
    \end{tabular}
    \caption{\textbf{Qualitative visualizations on Seen Split} showing three examples (rows) comparing image, PredCls, SGCls, and SGDet output graphs. Solid green lines represent accurately predicted groundtruths while solid red lines represent missed predictions. Dashed green lines represent visually meaningful predictions yet unannotated whereas dashed red lines represent predicted edges which don't align with the visuals.}
    \label{fig:qual_seen}
\end{figure*}

\begin{figure*}
    \centering
    \begin{tabular}{cccc}
        \includegraphics[width=0.24\textwidth]{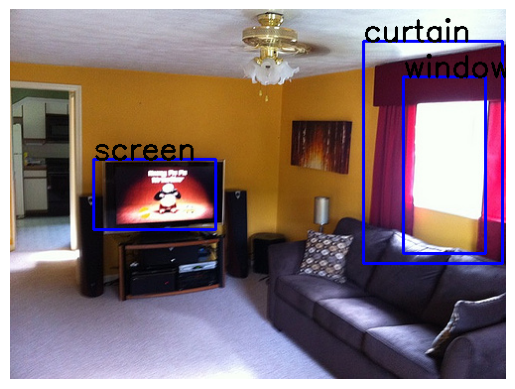} &  
        \includegraphics[width=0.24\textwidth]{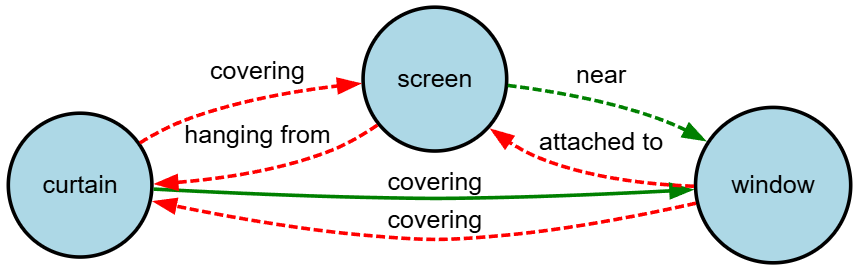} & 
        \includegraphics[width=0.24\textwidth]{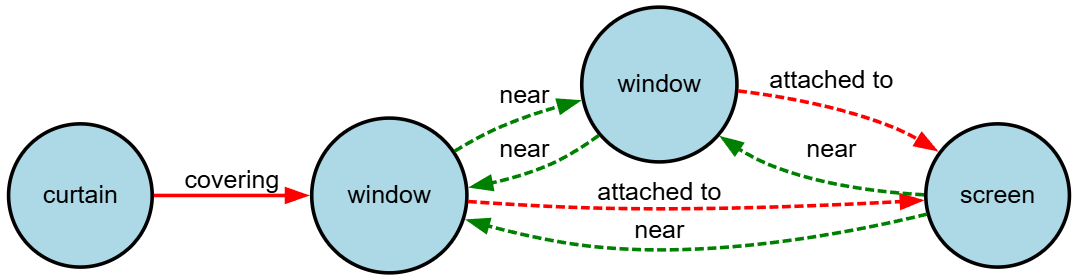} &
        \includegraphics[width=0.24\textwidth]{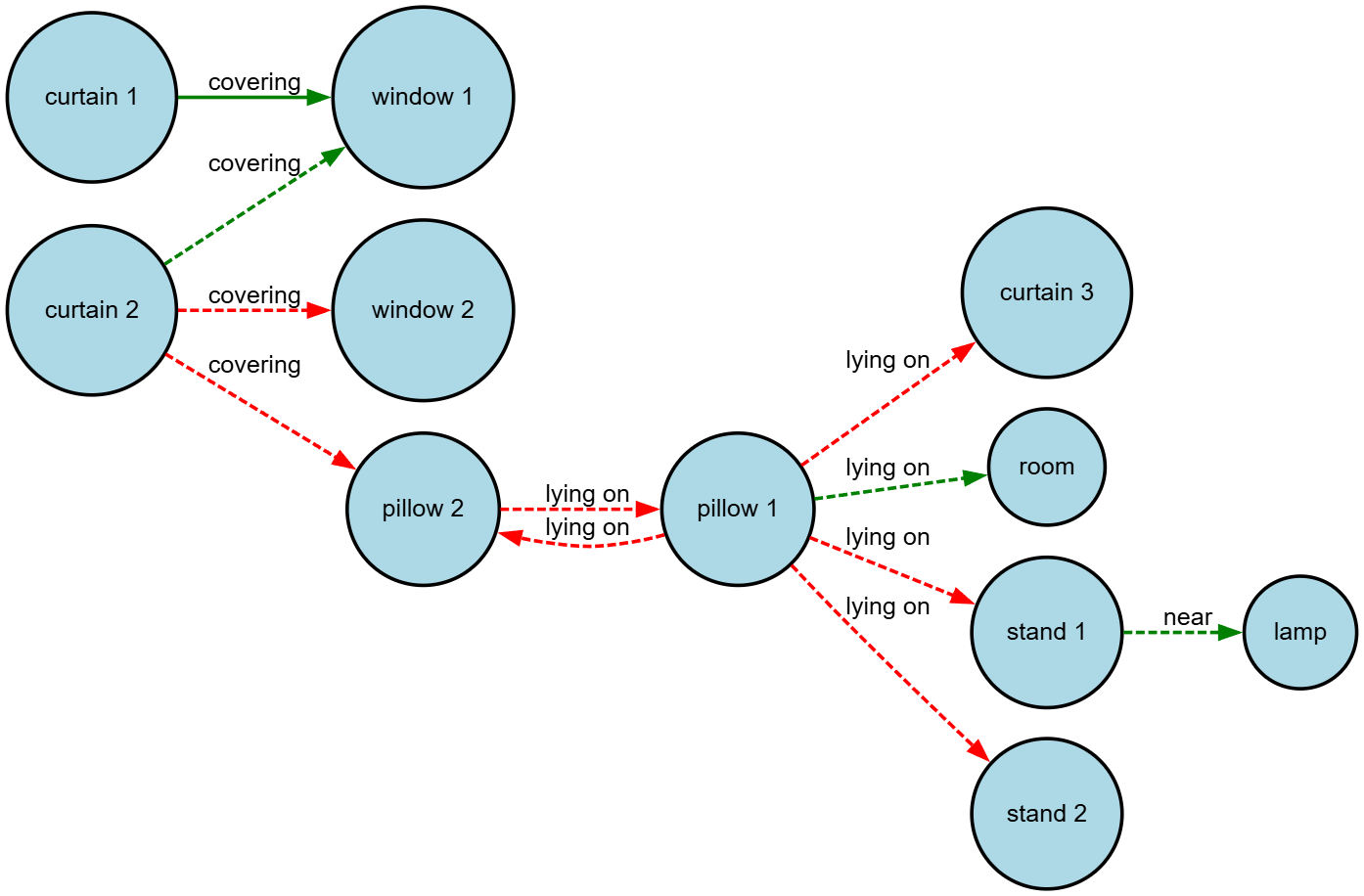} \\
        (a) & (b) & (c) & (d) \\[0.8em]
        
        \includegraphics[width=0.24\textwidth]{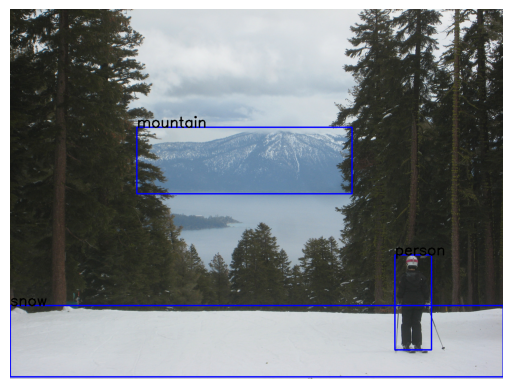} &  
        \includegraphics[width=0.24\textwidth]{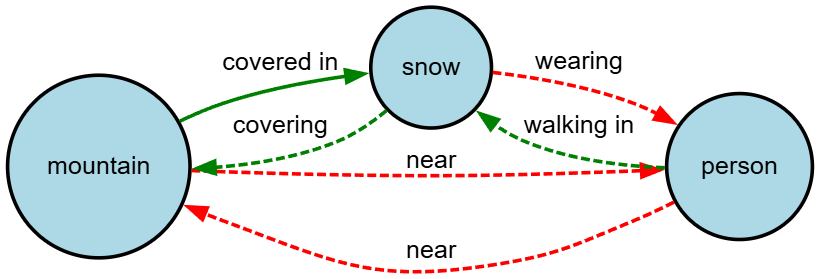} & 
        \includegraphics[width=0.24\textwidth]{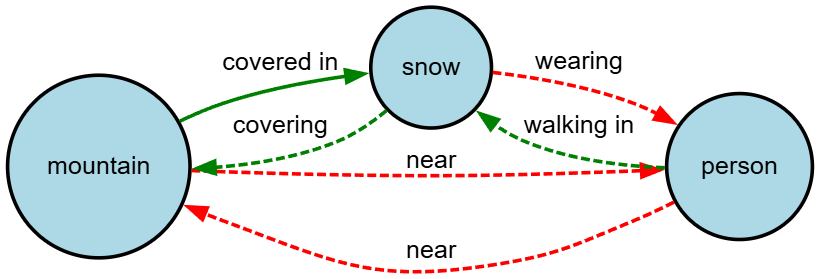} &
        \includegraphics[height=2.5cm]{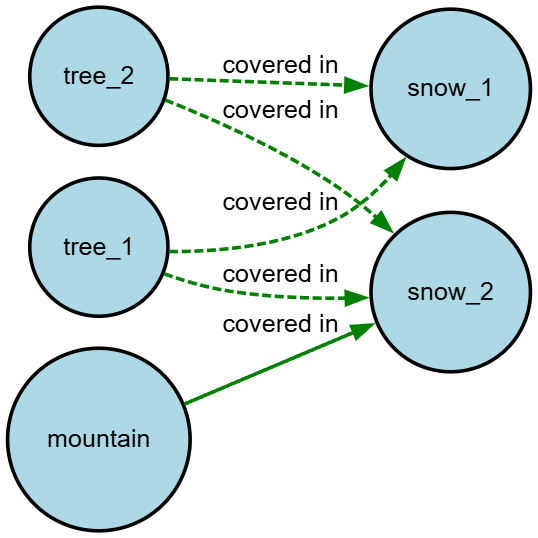} \\
        (e) & (f) & (g) & (h) \\[0.8em]
        
        \includegraphics[height=3cm]{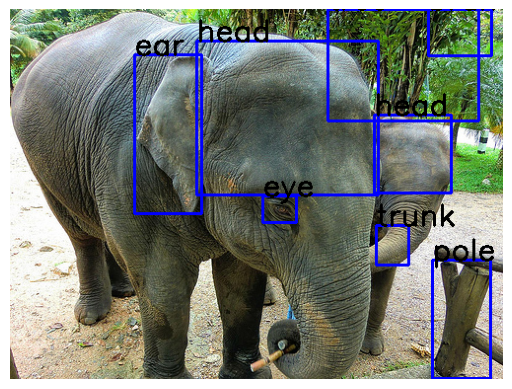} &  
        \includegraphics[width=0.24\textwidth]{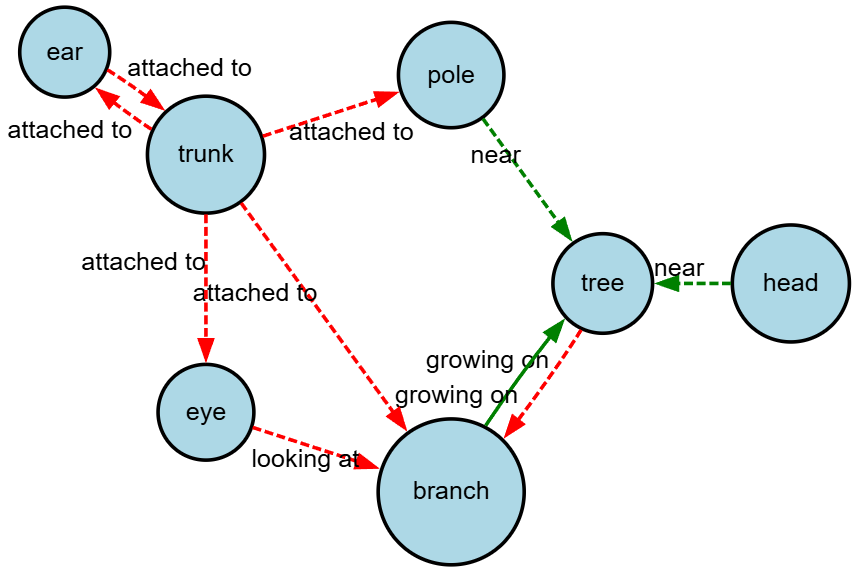} & 
        \includegraphics[width=0.24\textwidth]{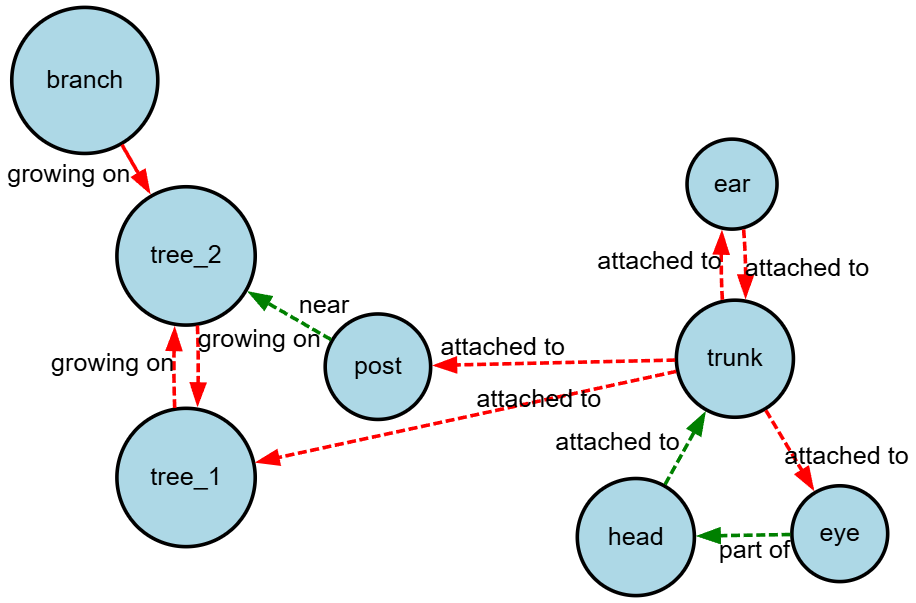} &
        \includegraphics[height=3cm]{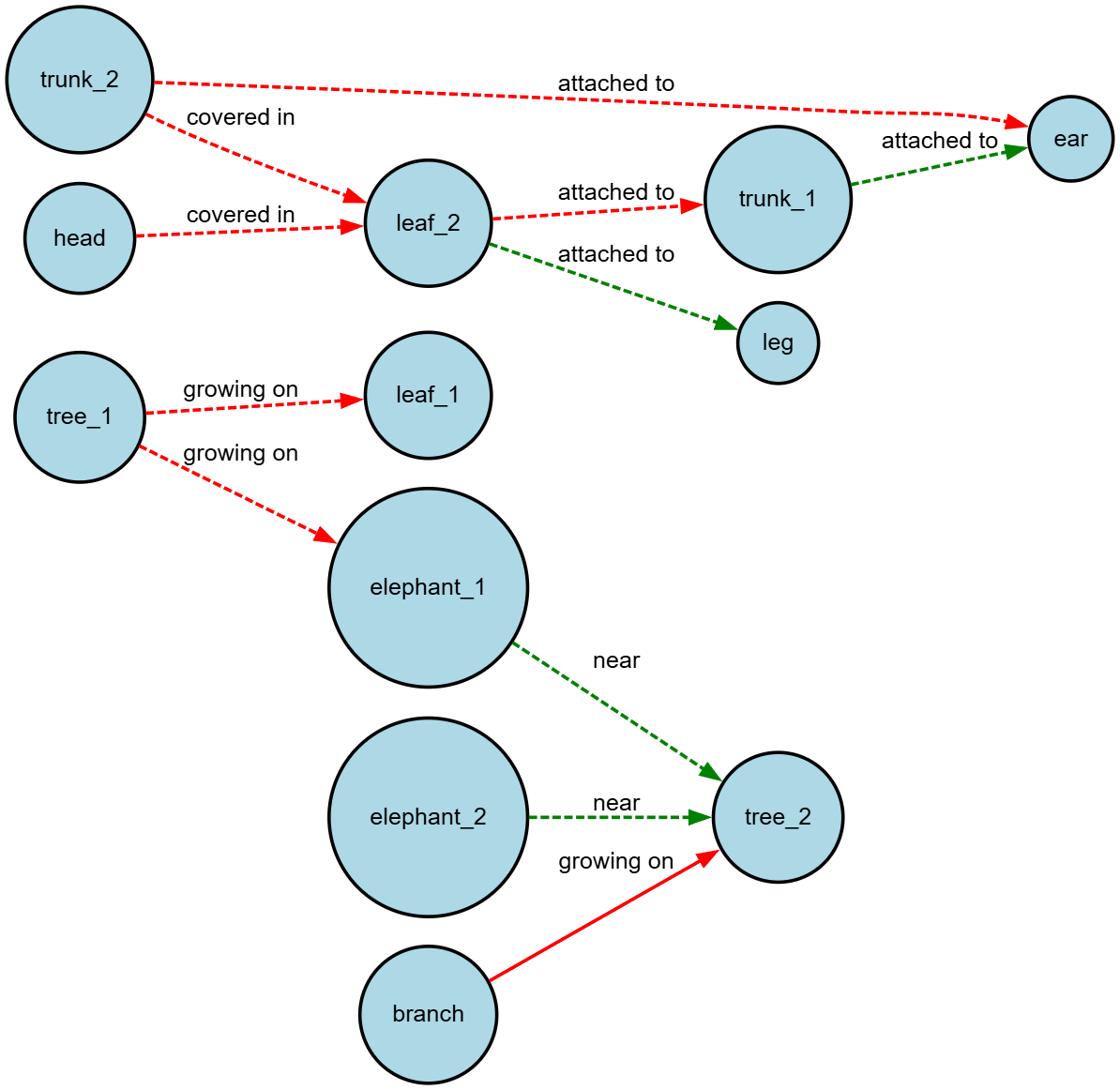} \\
        (i) & (j) & (k) & (l) \\
    \end{tabular}
    \caption{\textbf{Qualitative Visualizations on Unseen Split} showing three examples (rows) comparing image, PredCls, SGCls, and SGDet output graphs. Solid green lines represent accurately predicted groundtruths while solid red lines represent missed predictions. Dashed green lines represent visually meaningful predictions yet unannotated whereas dashed red lines represent predicted edges which don't align with the visuals.}
    \label{fig:qual_unseen}
\end{figure*}

\begin{figure}
    \centering
    \begin{tabular}{cccc}
        \includegraphics[width=0.24\textwidth]{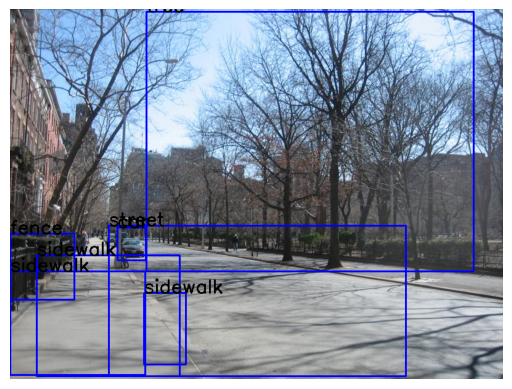} &  
        \includegraphics[width=0.24\textwidth]{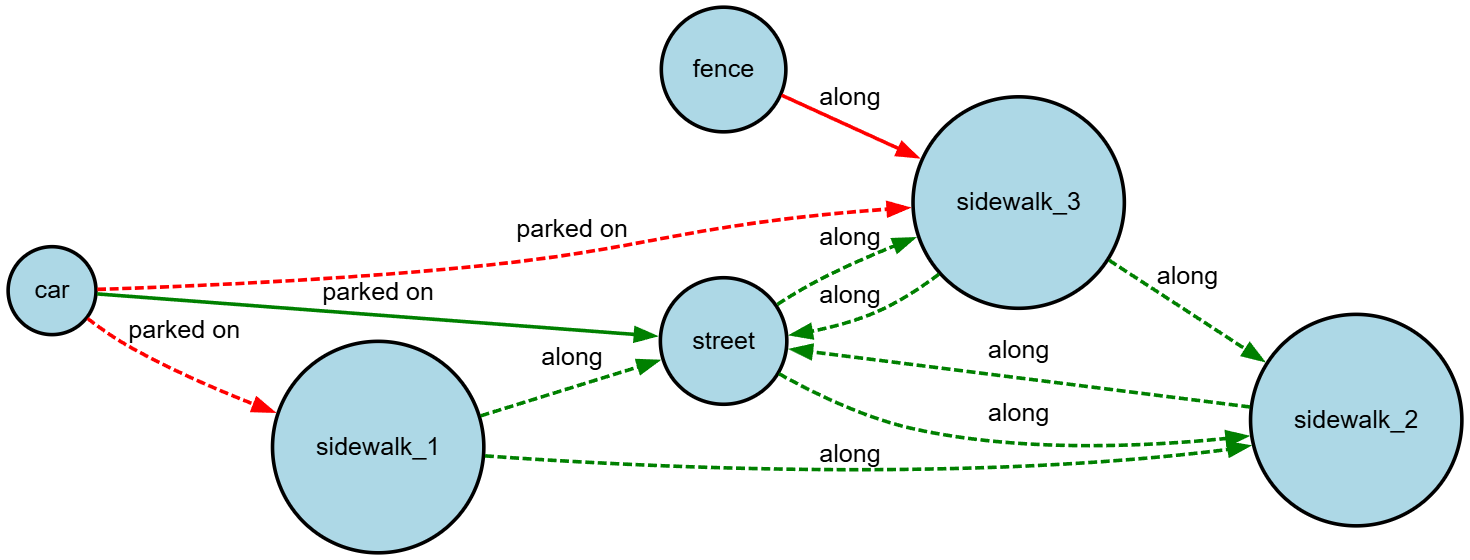} & 
        \includegraphics[width=0.24\textwidth]{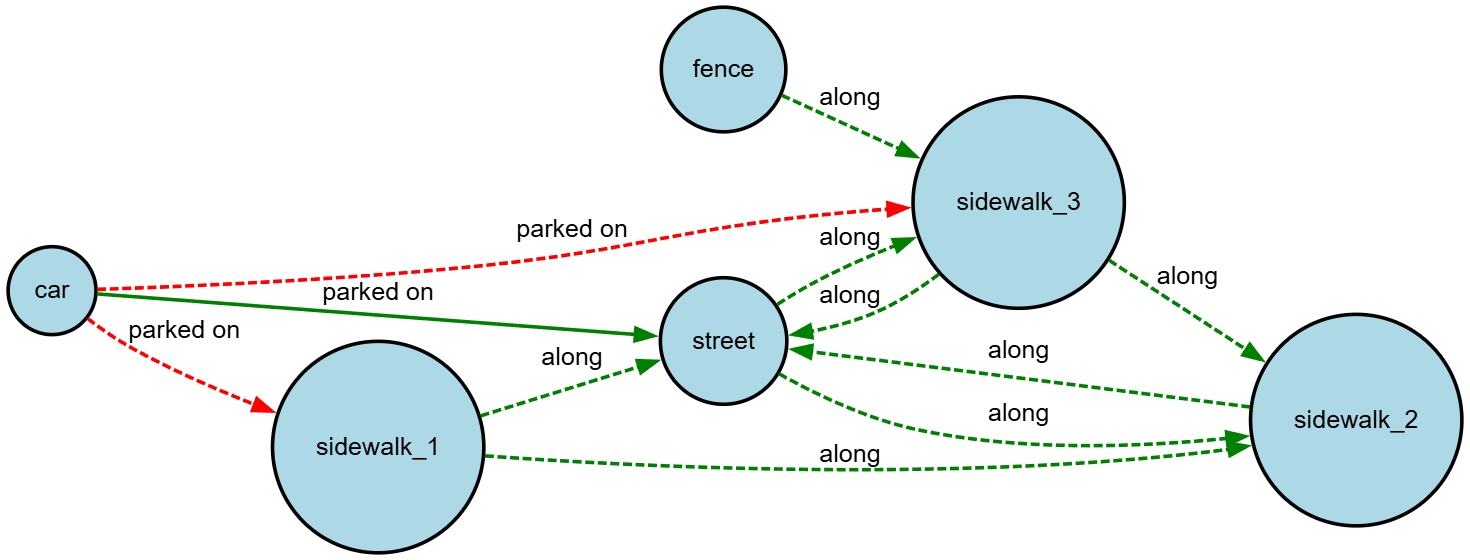} &
        \includegraphics[width=0.24\textwidth]{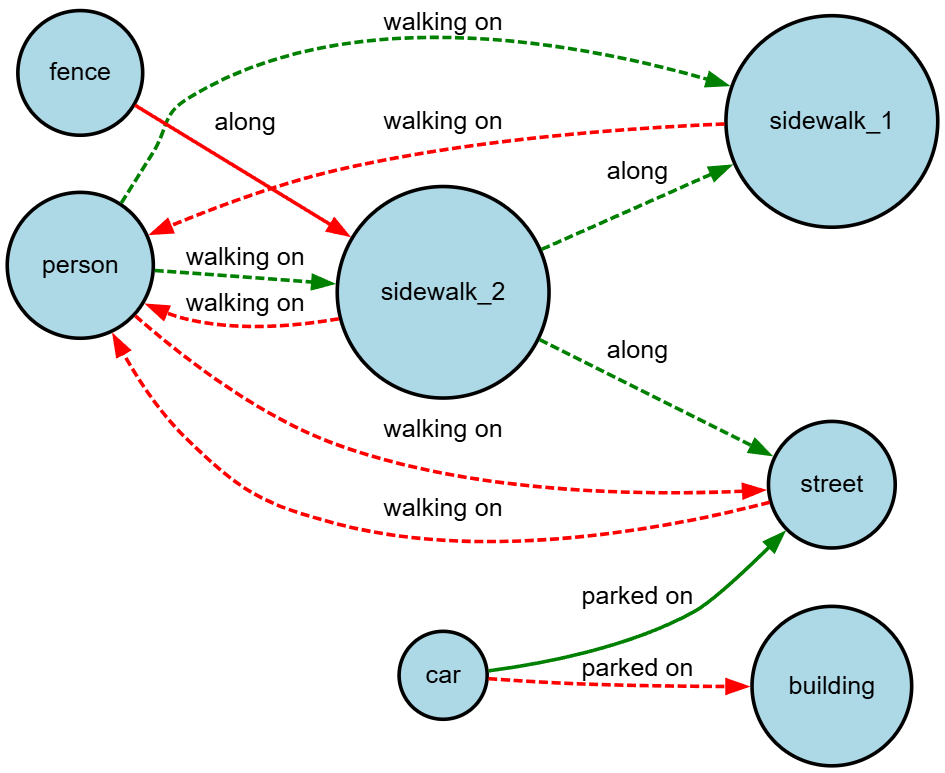} \\
        (a) & (b) & (c) & (d) \\[0.8em]
        
        \includegraphics[height=2.5cm]{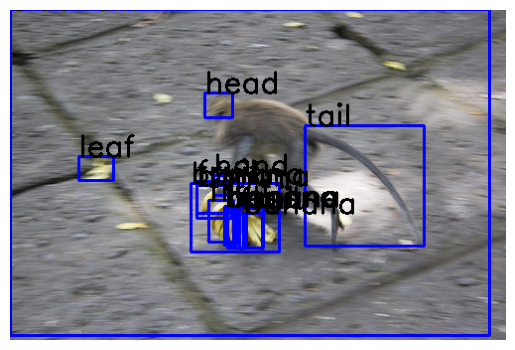} &  
        \includegraphics[width=0.24\textwidth]{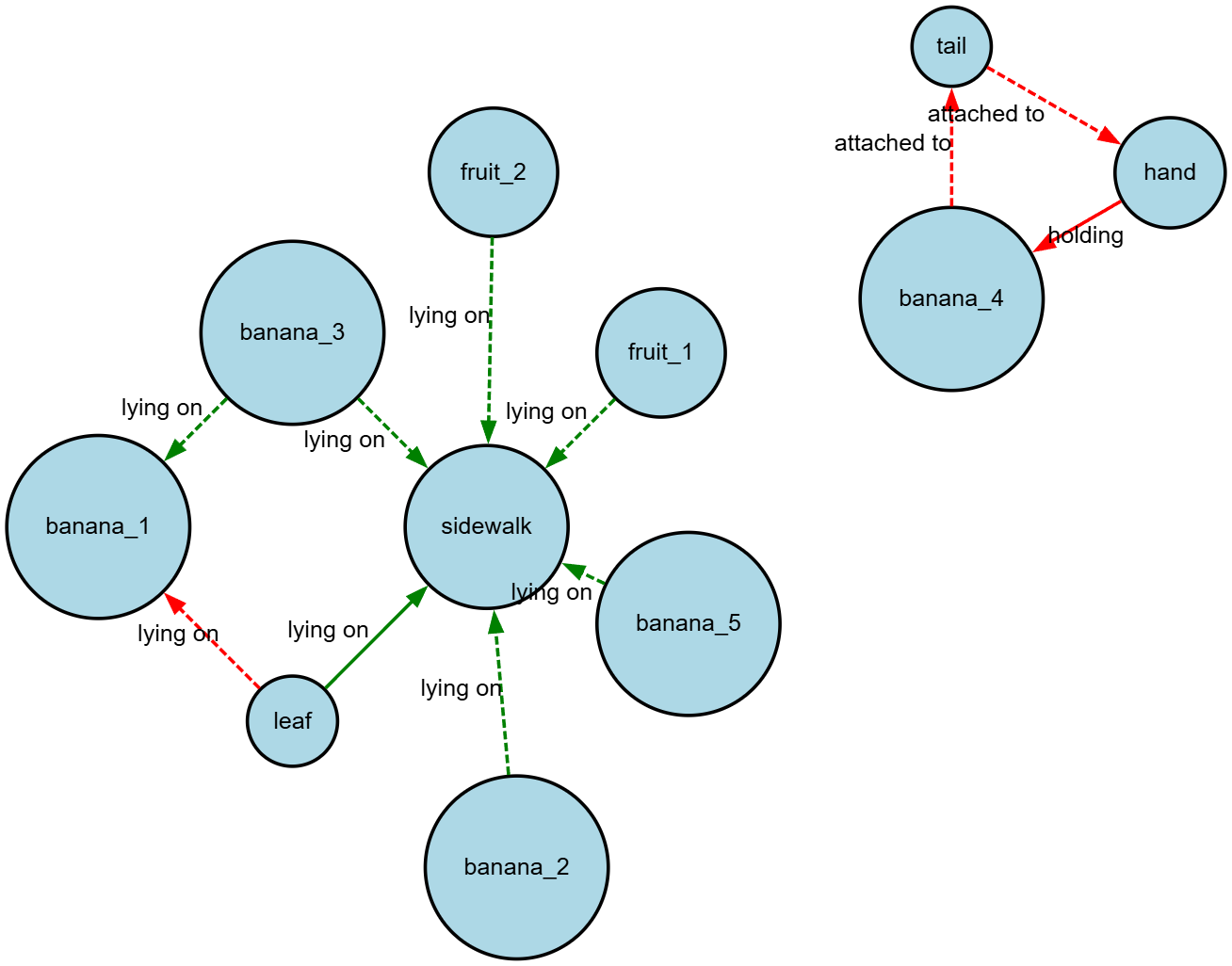} & 
        \includegraphics[height=3cm]{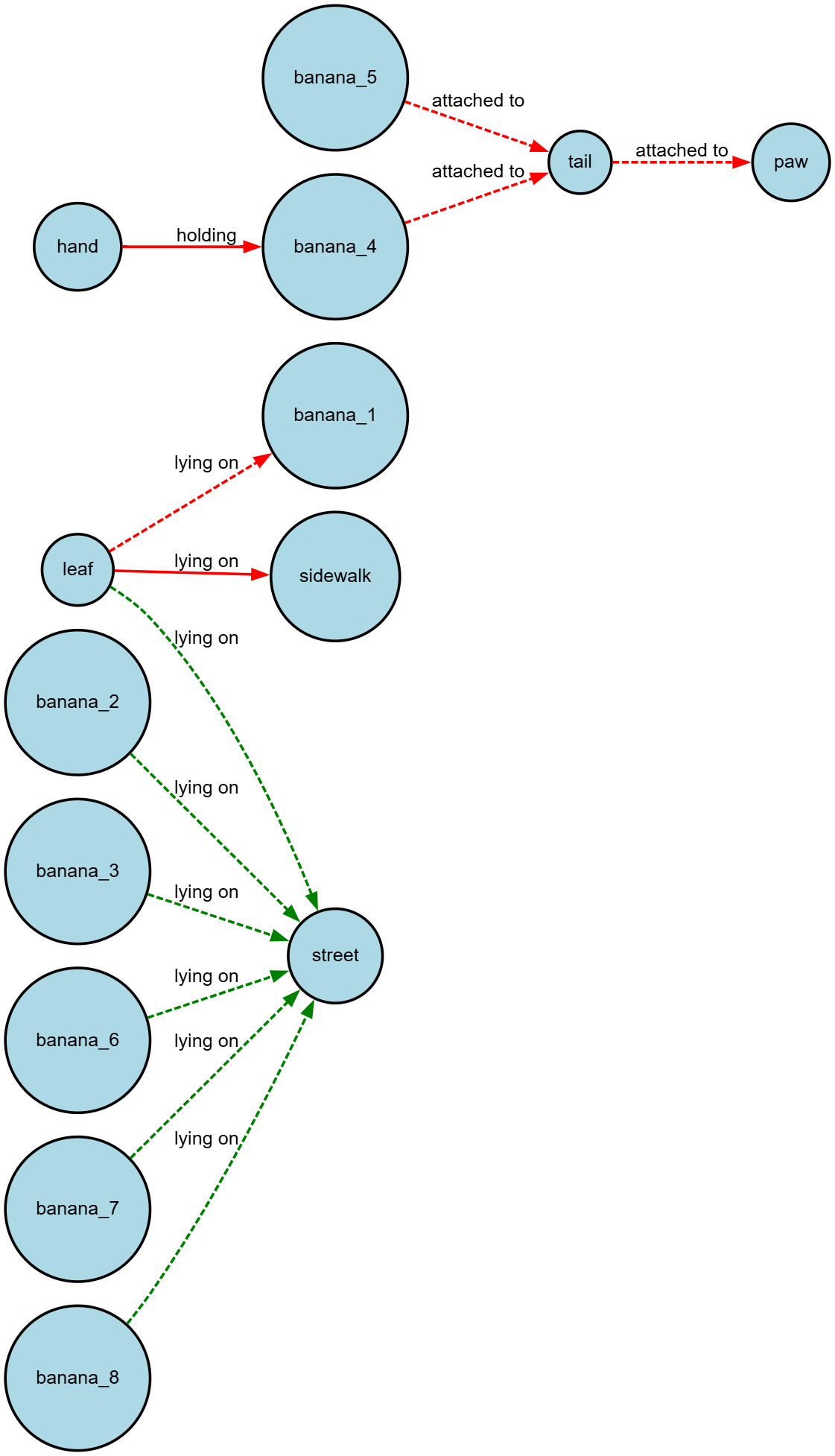} &
        \includegraphics[width=0.24\textwidth]{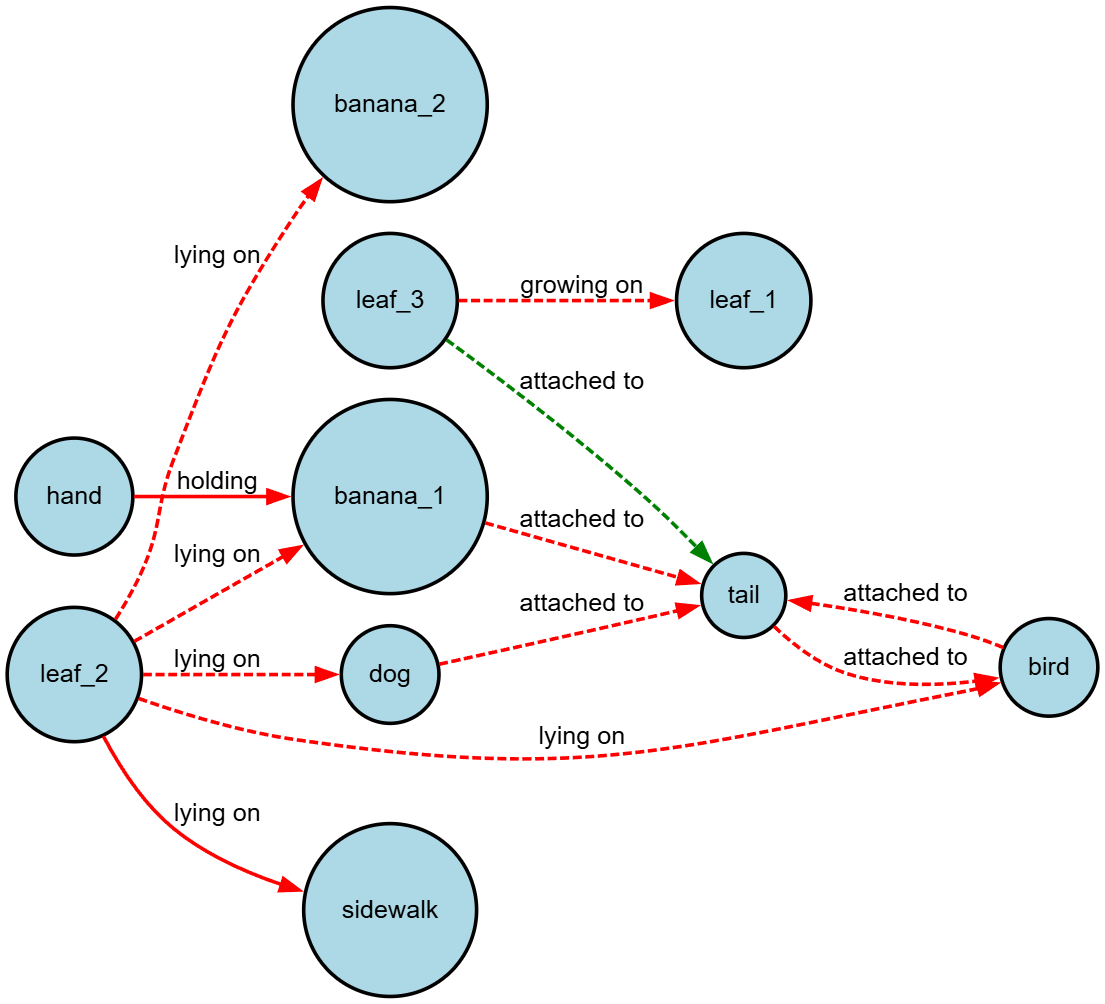} \\
        (e) & (f) & (g) & (h) \\[0.8em]
        
        \includegraphics[height=3cm]{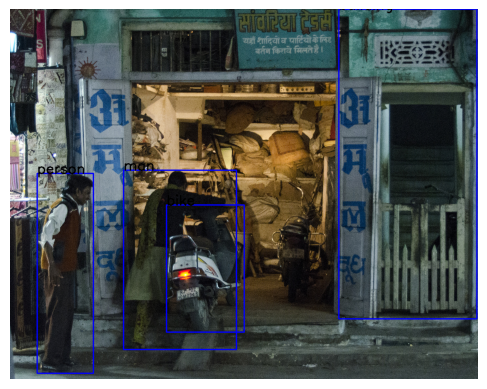} &  
        \includegraphics[width=0.24\textwidth]{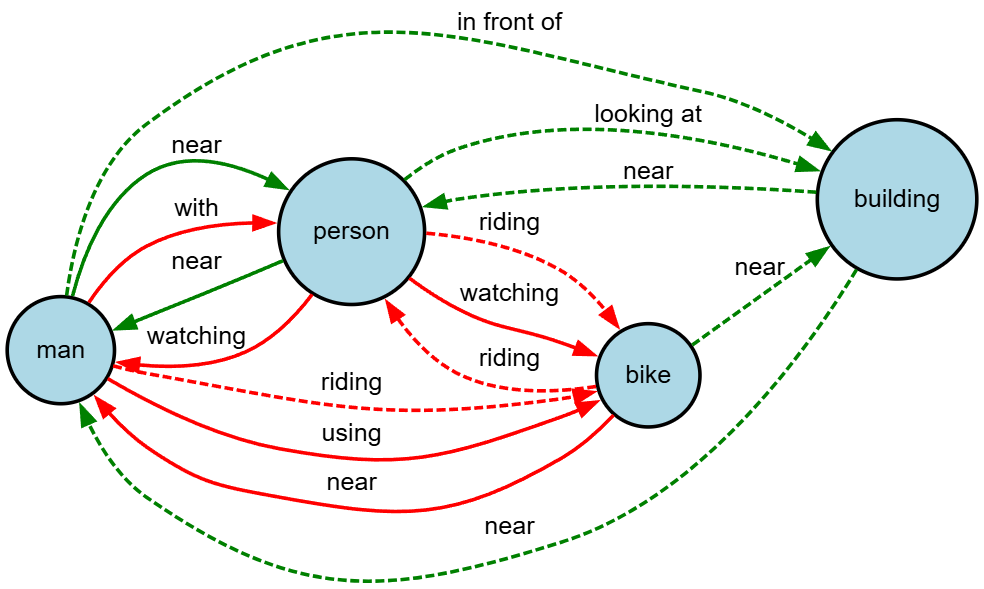} & 
        \includegraphics[width=0.24\textwidth]{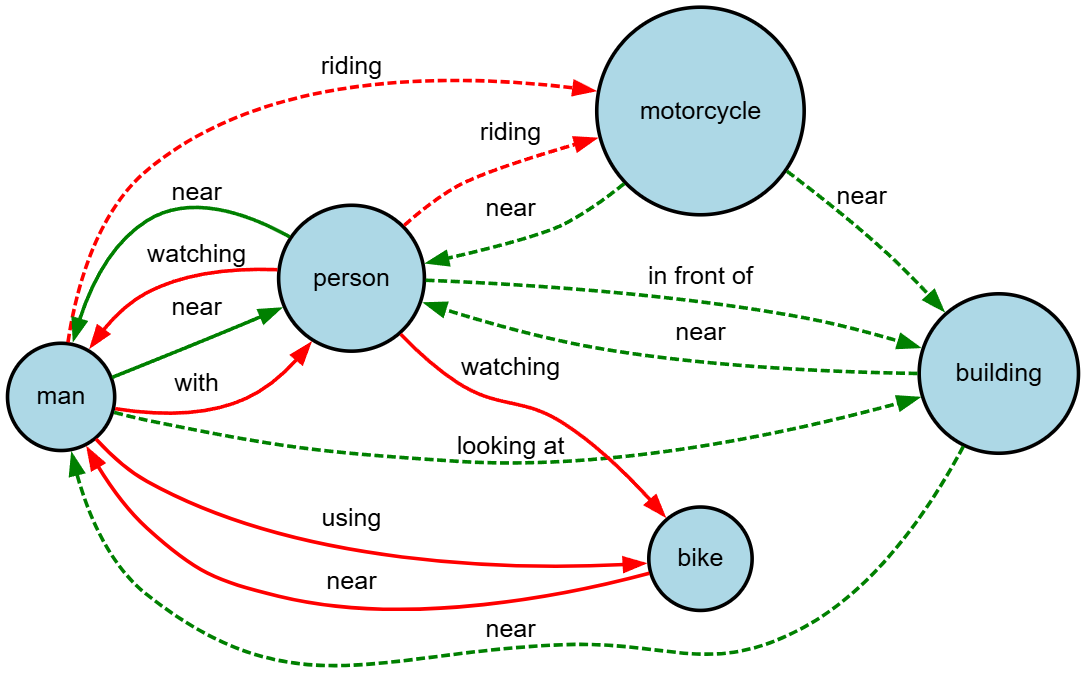} &
        \includegraphics[height=3cm]{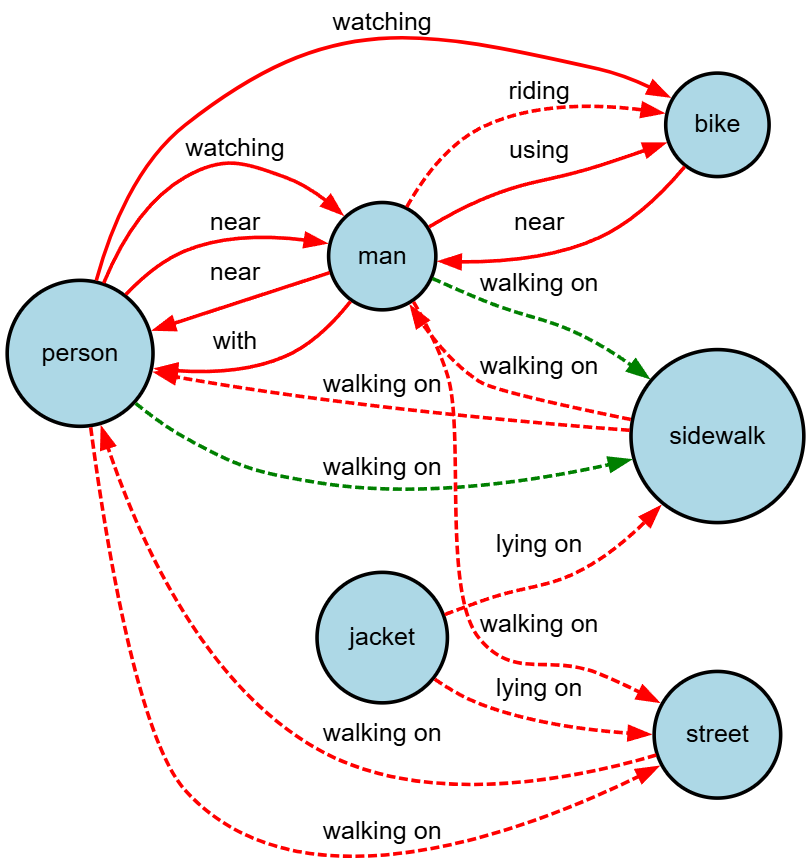} \\
        (i) & (j) & (k) & (l) \\
    \end{tabular}
    \caption{\textbf{Qualitative visualizations on Mixed Split} showing three examples (rows) comparing image, PredCls, SGCls, and SGDet output graphs. Solid green lines represent accurately predicted groundtruths while solid red lines represent missed predictions. Dashed green lines represent visually meaningful predictions yet unannotated whereas dashed red lines represent predicted edges which don't align with the visuals.}
    \label{fig:qual_mixed}
\end{figure}

We provide additional qualitative examples for all three tasks - PredCls, SGCls and SGDet, across all three splits in Figure ~\ref{fig:qual_seen}, ~\ref{fig:qual_unseen}, ~\ref{fig:qual_mixed}. The visualizations show that the model is able able to generalize well to both seen and unseen predicates most of the times. Although the model occasionally misses ground-truth edges—particularly as evaluation difficulty increases—it consistently predicts visually meaningful yet unannotated relationships, demonstrating the grounding capability of our approach.

\section{Broader Impacts}
This work proposes a scalable, annotation-efficient approach to visual relationship detection by leveraging language models as symbolic priors. It can potentially democratize structured scene understanding in low-resource settings and reduce reliance on costly human annotations. However, as our method inherits biases from vision and language models, care must be taken to ensure fairness and avoid reinforcing spurious or culturally-specific associations in downstream applications.

\section{Conclusion}
This supplementary highlights additional analyses and ablations supporting EM-Grounding, our proposed framework for generalized visual relationship detection. By grounding symbolic priors from LLMs via iterative refinement, EM-Grounding enables strong generalization even with limited supervision. While our results show consistent improvements across seen, unseen, and mixed predicate settings, the framework assumes access to reliable object detections and is currently limited to a fixed label space. Future directions include incorporating spatial context into the symbolic prior, extending to open-vocabulary setups, and introducing structured uncertainty into the refinement loop. These enhancements aim to further strengthen EM-Grounding’s applicability to real-world open-scene understanding.

\end{document}